\documentclass{article}

\usepackage[final]{neurips_2022}

\usepackage[utf8]{inputenc} %
\usepackage[T1]{fontenc}    %
\usepackage{hyperref}       %
\hypersetup{
    colorlinks = true,
    linkcolor = blue,
    anchorcolor = blue,
    citecolor = blue,
    filecolor = blue,
    urlcolor = blue
}
\usepackage{url}            %
\usepackage{booktabs}       %
\usepackage{amsfonts}       %
\usepackage{nicefrac}       %
\usepackage{microtype}      %
\usepackage{graphicx}
\usepackage{natbib}
\usepackage{doi}
\usepackage{xcolor}
\usepackage{amsmath,amsthm,amssymb,amsfonts}
\usepackage{bbm}
\usepackage{multirow}
\usepackage{multicol}
\usepackage{caption}
\usepackage{subcaption}
\usepackage{xspace}
\usepackage[normalem]{ulem}
\usepackage{tablefootnote}
\usepackage{wrapfig}
\usepackage{enumitem}

\DeclareMathOperator*{\argmin}{arg\,min}

\DeclareMathOperator{\EX}{\mathbb{E}}%

\newcommand{\methodacro}{SMaT\xspace}

\newcommand*\samethanks[1][\value{footnote}]{\footnotemark[#1]}

\title{Learning to Scaffold:\\
Optimizing Model Explanations for Teaching}

\author{
Patrick Fernandes\thanks{\,\, Equal contribution. Corresp. to \texttt{\small pfernand@cs.cmu.edu} or \texttt{\small marcos.treviso@tecnico.ulisboa.pt}} $^{,\Psi,\Omega,\Re}$ \qquad
Marcos Treviso\samethanks \phantom{  }$^{,\Omega,\Re}$ \qquad 
Danish Pruthi\thanks{\,\, Work done while at Carnegie Mellon University, prior to joining Amazon.}\phantom{  }$^{,\Lambda}$\\
\textbf{André F. T. Martins}$^{\Omega,\Re,\Gamma}$ \qquad
\textbf{Graham Neubig}$^{\Psi}$ \vspace{0.1cm}\\
$^\Psi$Language Technologies Institute, Carnegie Mellon University, Pittsburgh, PA \\
$^\Omega$Instituto Superior Técnico \& LUMLIS (Lisbon ELLIS Unit), Lisbon, Portugal \\
$^\Re$Instituto de Telecomunicações, Lisbon, Portugal  \\
$^\Lambda$Amazon Web Services \hspace{0.5cm} $^\Gamma$Unbabel, Lisbon, Portugal \\
\vspace{-0.5cm}
}
\date{}

\hypersetup{
pdftitle={Learning to Scaffold: Optimizing Model Explanations for Teaching},
pdfsubject={cs.LG, stat.ML},
pdfauthor={Patrick ~Fernandes, Marcos ~Treviso, ~Danish Pruthi, ~André F. T. Martins, ~Graham Neubig},
}

\begin{document}
\maketitle

\begin{abstract}
Modern machine learning models are opaque, and as a result there is a burgeoning academic subfield on methods that \emph{explain} these models' behavior. %
However, what is the precise goal of providing such explanations, and how can we demonstrate that explanations achieve this goal?
Some research argues that explanations should help \textit{teach} a student (either human or machine) to simulate the model being explained, and that the quality of explanations can be measured by the simulation accuracy of students on unexplained examples.
In this work, leveraging meta-learning techniques, we extend this idea to \emph{improve the quality of the explanations themselves}, specifically by optimizing explanations such that student models more effectively learn to simulate the original model.
We train models on three natural language processing and computer vision tasks, and find that students trained with explanations extracted with our framework are able to simulate the teacher significantly more effectively than ones produced with previous methods.
Through human annotations and a user study, we further find that these learned explanations more closely align with how humans would explain the required decisions in these tasks.
Our code is available at 
\href{https://github.com/coderpat/learning-scaffold}{https://github.com/coderpat/learning-scaffold}.
\end{abstract}

\section{Introduction}

While deep learning's performance has led it to become the dominant paradigm in machine learning, its relative opaqueness has brought great interest in methods to improve \emph{model interpretability}. 
Many recent works propose methods for extracting \emph{explanations} from neural networks (\autoref{sec:related-work}), which vary from the highlighting of relevant input features \citep{Simonyan2014DeepIC,arras-etal-2017-explaining,ding-etal-2019-saliency} to more complex representations of the reasoning of the network \citep{Mu2020CompositionalEO,wu-etal-2021-polyjuice}.
However, are these methods actually achieving their goal of making models more interpretable?
Some concerning findings have cast doubt on this proposition; different explanations methods have been found to disagree on the same model/input \citep{neely2021order,bastings2021-shortcuts} and explanations do not necessarily help predict a model's output and/or its failures \citep{chandrasekaran-etal-2018-explanations}.
\looseness=-1

In fact, the research community is still in the process of understanding \emph{what} explanations are supposed to achieve, and \emph{how} to assess success of an explanation method \citep{doshi2017towards,miller2019explanation}. 
Many early works on model interpretability designed their methods around a set of desiderata
\citep{pmlr-v70-sundararajan17a,lertvittayakumjorn-toni-2019-human}
and relied on qualitative assessment of a handful of samples with respect to these desiderata; a process that is highly subjective and is hard to reproduce.
In contrast, recent works have focused on more quantitative criteria: 
correlation between explainability methods for measuring \emph{consistency}~\citep{jain-wallace-2019-attention,serrano-smith-2019-attention}, 
\textit{sufficiency} and \textit{comprehensiveness}~\citep{deyoung-etal-2020-eraser}, and \emph{simulability}: whether a human or machine consumer
of explanations understands the model behavior well enough to predict its output on unseen examples~\citep{lipton-2018-mythos, doshi2017towards}.
Simulability, in particular, has a number of desirable properties, such as being intuitively aligned with the goal of \emph{communicating} the underlying model behavior to humans
and being measurable in manual and automated experiments \citep{treviso-martins-2020-explanation,hase-bansal-2020-evaluating,pruthi2020-evaluating}.

For instance, \citet{pruthi2020-evaluating} proposed a framework for automatic evaluation of simulability that, given a \emph{teacher model} and explanations of this model's predictions, trains a \emph{student model} to match the teacher's predictions.
The explanations are then evaluated with respect to how well they help a student \emph{learn to simulate} the teacher (\autoref{sec:background}).
This is analogous to the concept in pedagogy of \textbf{instructional scaffolding} \citep{van2010scaffolding}, a process through which a teacher adds support for students to aid learning.
More effective scaffolding---in our case, better explanations---is assumed to lead to better student learning.
However, while this previous work provides an attractive way to \emph{evaluate} existing explanation methods, it stops short of proposing a method to actually \emph{improve} them.

\begin{figure}
    \centering
    \includegraphics[width=0.95\textwidth]{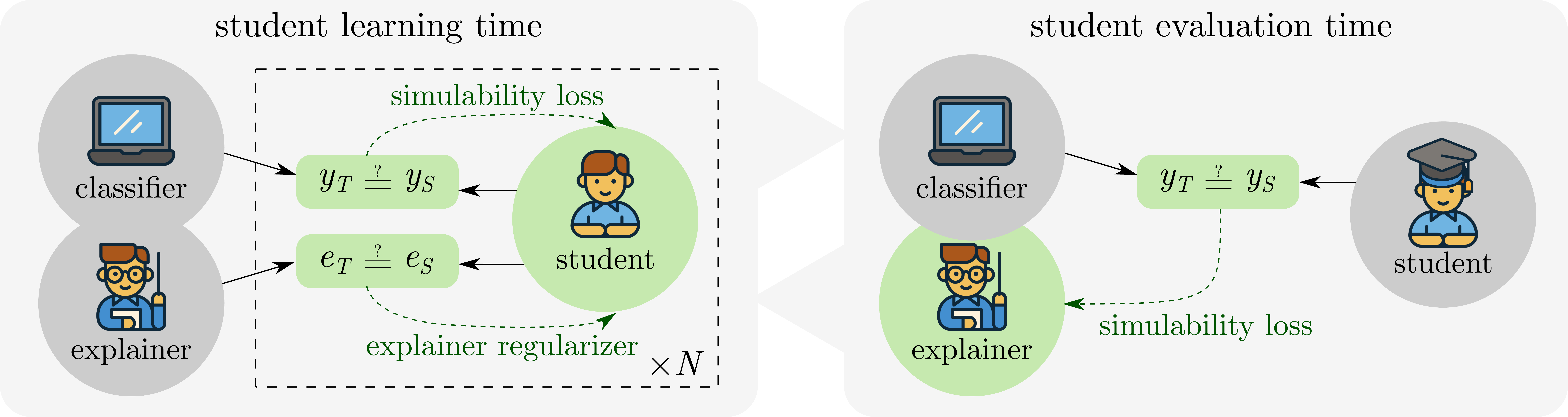}
    \caption{Illustration of our \methodacro framework. First, a student model is trained to recover the classifier's predictions and to match the explanations given by the explainer. Then,  the explainer is updated based on how well the trained student \textit{simulates} the classifier (without access to explanations). In practice, we repeat these two consecutive processes for several steps. \textcolor{green!60!black}{Green} arrows and boxes represent learnable components.
    }
    \label{fig:smat-framwork}
    \vspace{-1em}
\end{figure}

In this work, we propose to \emph{learn to explain} by directly learning explanations that provide better scaffolding of the student's learning, a framework we term \textit{\textbf{S}caffold-\textbf{Ma}ximizing \textbf{T}raining} (\textbf{\methodacro}).
\autoref{fig:smat-framwork} illustrates the framework: the explainer is used to \textit{scaffold} the student training, and is updated based on how well the student does at \textit{test} time at simulating the teacher model.
We take insights from research on meta-learning \citep{pmlr-v70-finn17a,raghu2020teaching}, 
formalizing our setting as a bi-level optimization problem and optimizing it based on higher-order differentiation (\autoref{sec:learn-explainers}). Importantly, our high-level framework makes few assumptions about the model we are trying to explain, the structure of the explanations or the modalities considered. 
To test our framework, we then introduce a \emph{parameterized} attention-based explainer optimizable with \methodacro that works for any model with attention mechanisms (\autoref{sec:attention-explainer}).

We experiment with \methodacro in text classification, image classification, and (multilingual) text-based regression tasks using pretrained transformer models (\autoref{sec:experiments}). We find that our framework is able to effectively optimize explainers across all the considered tasks, where students trained with \emph{learned} attention explanations achieve better simulability than baselines trained with \textit{static} attention or gradient-based explanations. We further evaluate the \emph{plausability} of our explanations (i.e., whether produced explanations align with how people would justify a similar choice) 
using human-labeled explanations (text classification and text regression) and through a human study (image classification) and find that explanations learned with SMaT are more plausible than the static explainers considered.
Overall, the results reinforce the utility of scaffolding as a criterion for evaluating and improving model explanations.

\section{Background}
\label{sec:background}

Consider a model $T: \mathcal{X} \rightarrow \mathcal{Y}$ trained on some dataset $\mathcal{D}_{\text{train}} = \{(x_i, y_i)\}_{i=1}^{N}$. For example, this could be a text or image classifier that was trained on a particular downstream task (with $\mathcal{D}_{\text{train}}$ being the training data for that task). \textit{Post-hoc} interpretability methods typically introduce an \textit{explainer} module $E_T: \mathcal{T} \times \mathcal{X} \rightarrow \mathcal{E}$ that takes a model and an input, and produces an explanation $e \in \mathcal{E}$ for the output of the model given that input, where $\mathcal{E}$ denotes the space of possible explanations. For instance, interpretability methods using saliency maps define $\mathcal{E}$ as the space of \textit{normalized} distributions of importance over $L$ input elements $e \in \triangle_{L-1}$ (where $\triangle_{L-1}$ is the $(L-1)$-probability simplex).

\cite{pruthi2020-evaluating} proposed an automatic framework for evaluating explainers that trains a \emph{student} model $S_{\theta}: \mathcal{X} \rightarrow \mathcal{Y}$ with parameters $\theta$ to \emph{simulate} the \emph{teacher} (i.e., the original classifier) in a \emph{constrained} setting. For example, the student can be constrained to have less capacity than the teacher by using a simpler model or trained with a subset of the dataset used for the teacher ($\hat{\mathcal{D}}_{\text{train}} \subsetneq \mathcal{D}_{\text{train}}$). 

In this framework, a baseline student $S_\theta$ is trained according to $\theta^* = \argmin_{\theta} \EX_{(x, y) \sim \hat{\mathcal{D}}_{\text{train}}} \left [ \mathcal{L}_\mathrm{sim}(S_{\theta}(x), T(x)) \right ]$, and its simulability $\textsc{sim}(S_{\theta^*}, T)$ is measured on an unseen test set.
The actual form of $\mathcal{L}_\mathrm{sim}$ and $\textsc{sim}(S_{\theta^*}, T)$ is task-specific. For example, in a classification task, we use cross-entropy as the simulation loss $\mathcal{L}_\mathrm{sim}$ over the teacher's predictions, while the simulability of a model $S_{\theta^*}$ can be defined as the simulation accuracy, i.e., what percentage of the student and teacher predictions match over a \textit{held-out} test set $\mathcal{D}_{\text{test}}$:
\begin{equation}
\textsc{sim}(S_{\theta^*}, T) = \EX_{(x, y)\sim \mathcal{D}_{\text{test}}} [\mathbbm{1}\{S_{\theta^*}(x) = T(x)\}].
\label{eq:sim}
\end{equation}
Next, the training of the student is augmented with explanations produced by the explainer $E$. We introduce a student explainer $E_S: \mathcal{S} \times \mathcal{X} \rightarrow \mathcal{E}$, (the $S$-explainer) to extract explanations from the student, and \textit{regularizing} these explanations on the explanations of teacher (the $T$-explainer), using a loss $\mathcal{L}_\mathrm{expl}$ that takes explanations for both models:
\begin{equation}
\theta^*_E = \argmin_{\theta} \EX_{(x, y) \sim \hat{\mathcal{D}}_{\text{train}}} \biggl [ \underbrace{\mathcal{L}_\mathrm{sim} \left (S_{\theta}(x), T(x) \right )}_{\text{simulability loss}} + \beta \underbrace{\mathcal{L}_\mathrm{expl} \left (E_{S}(S_{\theta}, x), E_{T}(T, x) \right )}_\text{explainer regularizer} \biggr ].
\end{equation}

For example, \citet{pruthi2020-evaluating} considered as a teacher explainer $E_{T}$ various methods such as LIME \citep{lime}, Integrated Gradients \citep{pmlr-v70-sundararajan17a}, and attention mechanisms, and explored both attention regularization (using Kullback-Leibler divergence) and multi-task learning to regularize the student. 

The key assumption surrounding this evaluation framework is that a student trained with \emph{good} explanations should learn to simulate the teacher better than a student trained with bad or no explanations, that is, $\textsc{sim} \left (S_{\theta_E^*}, T \right ) > \textsc{sim}\left (S_{\theta^*}, T  \right ).$ For clarity, we will refer to the simulability of a model $S_{\theta_E^*}$ trained using explanations as \emph{scaffolded} simulability.

\section{Optimizing Explainers for Teaching}
\label{sec:learn-explainers}

As a \textbf{first contribution} of this work, we extend the previously described framework to make it possible to directly optimize the teacher explainer so that it can most effectively teach the student the original model's behavior. To this end, consider a \textit{parameterized} $T$-explainer $E_{\phi_T}$ with parameters $\phi_T$,
and equivalently a \textit{parameterized} $S$-explainer $E_{\phi_S}$ with parameters $\phi_S$. 
We can write the loss function for the student and $S$-explainer as:
\begin{align}
\mathcal{L}_\textrm{student}(S_{\theta}, E_{\phi_S}, T, E_{\phi_T}, x)  &= \mathcal{L}_\mathrm{sim} \left( S_{\theta}(x), T(x) \right) +  \beta \mathcal{L}_\mathrm{expl} \left (E_{\phi_S}(S_{\theta}, x), E_{\phi_T}(T, x) \right ). \label{eq:loss_student}
\end{align}

While this framework is flexible enough to rigorously and automatically evaluate many types of explanations, calculating scaffolded simulability requires an optimization procedure to learn the student and $S$-explainer parameters $\theta,\phi_S$.
This makes it non-trivial to achieve our goal of directly finding the teacher explainer parameters $\phi_T$ that optimize scaffolded simulability.
To overcome this challenge, we draw inspiration from the extensive literature on meta-learning \citep{schmidhuber:1987:srl, pmlr-v70-finn17a}, and frame the optimization as the following bi-level optimization problem (see \citet{grefenstette2019generalized} for a primer):
\begin{align}
\theta^*(\phi_T), \phi_S^*(\phi_T) &= \argmin_{\theta,\phi_S} \EX_{(x, y) \sim \hat{\mathcal{D}}_{\text{train}}} \left [ \mathcal{L}_\textrm{student}(S_{\theta}, E_{\phi_S}, T, E_{\phi_T}, x)  \right ] \label{eq:inner_opt} \\
\phi_{T}^* &= \argmin_{\phi_T} \EX_{(x, y) \sim \mathcal{D}_{\text{test}}} \left [ \mathcal{L}_\mathrm{sim} \left (S_{\theta^*(\phi_T)}(x), T(x) \right ) \right ]. \label{eq:outer_opt} 
\end{align}
Here, the \emph{inner} optimization updates the student and the $S$-explainer parameters (\autoref{eq:inner_opt}), and in the \emph{outer} optimization we update the $T$-explainer parameters (\autoref{eq:outer_opt}).
\textbf{Importantly}, our framework does not modify the teacher, as our goal is to explain a model without changing its original behavior. Notice that we also simplify the problem by considering the more tractable simulation loss $\mathcal{L}_\mathrm{sim}$ instead of the simulability metric $\textsc{sim}(S_{\theta^*}, T)$ as part of the objective for the outer optimization.

Now, if we assume the explainers $E_{\phi_T}$ and $E_{\phi_S}$ are differentiable, we can use gradient-based optimization \citep{pmlr-v70-finn17a} to optimize both the student (with its explainer) and the $T$-explainer. 
In particular, we use \emph{explicit} differentiation to solve this optimization problem.
To compute gradients for $\phi_T$, we have to differentiate through a gradient operation, which requires Hessian-vector products, an operation supported by most modern deep learning frameworks \citep{jax2018github, grefenstette2019generalized}. However, explicitly computing gradients for $\phi_T$ through a large number of inner optimization steps is computationally intractable. To circumvent this problem, typically the inner optimization is run for only a couple of steps or a \emph{truncated} gradient is computed \citep{pmlr-v89-shaban19a}. In this work, we take the approach of taking a \emph{single} inner optimization step and learning the student and $S$-explainer jointly with the $T$-explainer \textit{without} resetting the student~\citep{Dery2021ShouldWB}. At each step, we update the student and $S$-explainer parameters as follows: 
\begin{align}
\theta^{t+1} &= \theta^{t} - \eta_{\textsc{inn}} \nabla_{\theta}\EX_{(x, y) \sim \hat{\mathcal{D}}_{\text{train}}} \left [ \mathcal{L}_\textrm{student}(S_{\theta^{t}}, E_{\phi_S^{t}}, T, E_{\phi_T^{t}}, x)  \right ] \\
\phi_{S}^{t+1} &= \phi_{S}^{t} - \eta_{\textsc{inn}} \nabla_{\phi_S}\EX_{(x, y) \sim \hat{\mathcal{D}}_{\text{train}}} \left [ \mathcal{L}_\textrm{student}(S_{\theta^{t}}, E_{\phi_S^{t}}, T, E_{\phi_T^{t}}, x)  \right ].
\end{align}
After updating the student, we take an extra gradient step with the new parameters but only use these updates to calculate the \emph{outer}-gradient for $\phi_T$, without actually updating $\theta$. 
This approach is similar to the \textit{pilot update} proposed by \citet{zhou2021meta}, and we verified that it led to more stable optimization in practice: 
\begin{align}
\theta(\phi^t_T) = \theta^{t+1} - \eta_{\textsc{inn}} &\nabla_{\theta}\EX_{(x, y) \sim \hat{\mathcal{D}}_{\text{train}}} \left [ \mathcal{L}_\textrm{student}(S_{\theta^{t+1}}, E_{\phi_S^{t+1}}, T, E_{\phi_T^{t}}, x)  \right ] \\
\phi_{T}^{t+1} = \phi_{T}^{t} -  \eta_{\textsc{out}} &\nabla_{\phi_T}\EX_{(x, y) \sim \mathcal{D}_{\text{test}}} \left [ \mathcal{L}_\mathrm{sim} \left (S_{\theta(\phi_T^{t})}(x), T(x) \right ) \right ].
\end{align}

\section{Parameterized Attention Explainer}
\label{sec:attention-explainer}

As a \textbf{second contribution} of this work, we introduce a novel \emph{parameterized} attention-based explainer that can be learned with our framework. Transformer models \citep{NIPS2017_3f5ee243} are currently the most successful deep-learning architecture across a variety of tasks \citep{wikitext_sota,imagenet_sota}.
Underpinning their success is the \emph{multi-head attention mechanism}, which computes a \emph{normalized} distribution over the $1 \leq i \leq L$ input elements in parallel for each head $h$:
\begin{equation}
A^h = \textsc{softmax}(Q^h (K^h)^\top),
\end{equation}
where $Q^h = [q_0^h, \cdots, q_L^h]$ and $K^h = [k_0^h, \cdots, k_L^h]$ are the \emph{query} and \emph{key} linear projections over the input element representations for head $h$. Attention mechanisms have been used extensively for producing saliency maps \citep{wiegreffe-pinter-2019-attention, Vashishth2019AttentionIA} and while some concerns have been raised regarding their faithfulness \citep{jain-wallace-2019-attention}, overall attention-based explainers have been found to lead to relatively good explanations in terms of \emph{plausibility} and \emph{simulability} \citep{treviso-martins-2020-explanation,kobayashi-etal-2020-attention,pruthi2020-evaluating}.

However, to extract explanations from multi-head attention, we have two important design choices: 
\begin{enumerate}[leftmargin=*]
    \item \textbf{Single distribution selection:} Since self-attention produces an attention matrix $A^h \in \triangle^L_{L-1}$, we need to \emph{pool} these attention distributions to produce a single saliency map $e \in \triangle_{L-1}$. Typically, the distribution from a single token (such as \texttt{[CLS]}) or the \emph{average} of the attention distributions from all tokens $1 \leq i \leq L$ are used.
    \item \textbf{Head selection:} We also need to \emph{pool} the distributions produced by each head. Typical ad-hoc strategies include using the mean over all heads for a certain layer \citep{fomicheva2021translation} or selecting a single head based on plausibility on validation set \citep{treviso-etal-2021-ist}. However, since transformers can have hundreds or even thousands of heads, these choices rely on human intuition or require large amounts of plausibility labels.
\end{enumerate}

In this work, we approach the latter design choice in a more principled manner. Concretely, we associate each head with a weight and then perform a weighted sum over all heads.
These weights are learned such that the resulting explanation maximizes simulability, as described in \autoref{sec:learn-explainers}.
More formally, given a model $T_{\theta_T}$ and its query and key projections for an input $x$ for each layer and head $h \leq H$, we define a \emph{parameterized}, \emph{differentiable} attention explainer $E_{\phi_T}(T_{\theta_T}, x)$ as 
\begin{align}
    s^h = \frac{1}{L} \sum_{i=1}^L (q^h_i)^\top K^h, \qquad E_{\phi_T}(T, x) = \textsc{softmax} \left ( \sum_{h=1}^H \lambda_T^h s^h \right ),
\end{align}
where the teacher's head coefficients $\lambda_T \in \triangle_{H-1}$ are $\lambda_T = \textsc{normalize} (\phi_T)$  with $\phi_T \in \mathbb{R}^{H}$.

\begin{wrapfigure}{r}{8cm}
    \centering
    \includegraphics[height=3.5cm]{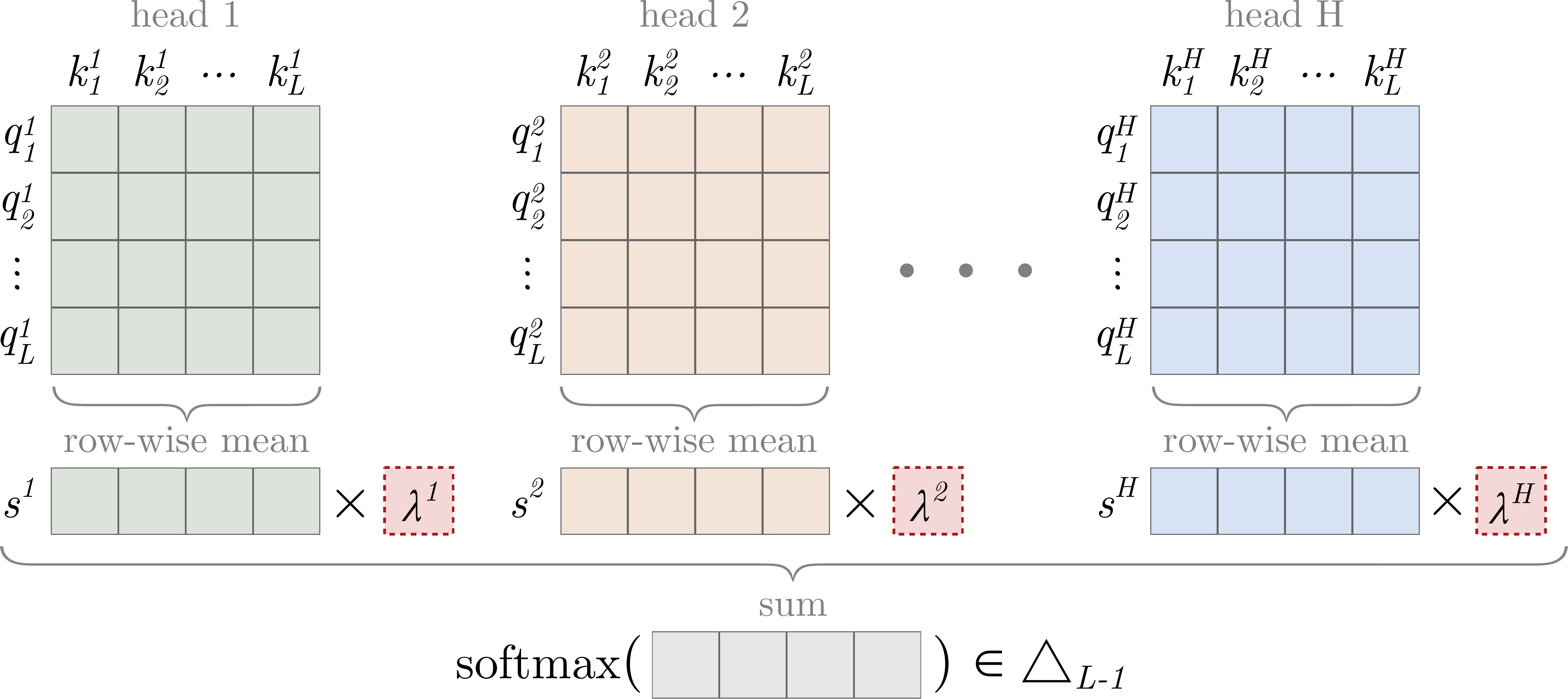}
    \caption{Our parameterized attention-based explainer. Dashed red boxes represent learned parameters $\lambda_T = \textsc{Sparsemax}(\phi_T) \in \triangle_{H-1}$, weighting average attention logits of each head $1 \leq h \leq H$. A softmax over the weighted sum generates the attention probabilities.}
    \label{fig:param_attention_steps}
\end{wrapfigure}

In this formulation, $s^h \in \mathbb{R}^{L}$ represents the average \emph{unnormalized attention logits} over all input elements, which are then combined according to $\lambda_T$ and normalized with \textsc{softmax} to produce a distribution in $\triangle_{L-1}$. We apply a normalization function $\textsc{normalize}$ to head coefficients involved to create a \emph{convex} combination over all heads in all layers. In this work we consider the sparse projection function $\textsc{normalize} =\textsc{Sparsemax}$ \citep{pmlr-v48-martins16}, as:
$$
\textsc{Sparsemax}(z) = \argmin_{p\in \triangle_{H-1}}\|p - z\|_2.
$$
We choose \textsc{Sparsemax} due to its benefits in terms of interpretability, since it leads to many heads having zero weight. We also found it outperformed every other projection we tried (see \autoref{app:importance-head-projection} for a more detailed discussion). \autoref{fig:param_attention_steps} illustrates each step of our parameterized attention explainer.

\section{Experiments}
\label{sec:experiments}

To evaluate our framework, we attempt to learn explainers for transformer models trained on three different tasks: text classification (\autoref{sec:exp:textclass}), image classification (\autoref{sec:exp:imageclass}), and machine translation quality estimation (a text-based regression task, detailed in \autoref{sec:exp:qe}). We use JAX \citep{jax2018github} to implement the higher-order differentiation, and use pretrained transformer models from the Huggingface Transformers library \citep{wolf-etal-2020-transformers}, together with Flax \citep{flax2020github}. 
For each task, we train a teacher model with AdamW \citep{loshchilov2018decoupled} but, as explained in \autoref{sec:learn-explainers}, we use SGD for the student model (inner loop). We also use scalar mixing \citep{peters-etal-2018-deep} to pool representations from different layers automatically.\footnote{While scalar mixing reduced variance of student performance, SMaT also worked with other common pooling methods.} We train students with a teacher explainer in three settings: 

\begin{itemize}[leftmargin=*]
\setlength\itemsep{-0.1em}
\item \textbf{No Explainer}: No explanations are provided, and no explanation regularization is used for training the student (i.e.~$\beta = 0$ in \autoref{eq:loss_student}).
We refer to studentsin this setting as \textbf{baseline} students. 
\item \textbf{Static Explainer}: Explanations for the teacher model are extracted with five commonly-used saliency-based explainers: 
(1) L2 norm of gradients; (2) a \textit{gradient $\times$ input} explainer \citep{Denil2014ExtractionOS}; 
(3) an \textit{integrated gradients} explainer \citep{pmlr-v70-sundararajan17a}; and \textit{attention} explainers that uses the \textit{mean} pooling over attention from 
(4) all heads in the model and 
(5) from the heads of the last layer \citep{fomicheva2021translation,vafa-etal-2021-rationales}.
More details can be found in \autoref{app:explainer-details}.
\item \textbf{Learned Explainer (SMaT)}: Explanations are extracted with the explainer described in \autoref{sec:attention-explainer}, with coefficients for each head that are trained with \textbf{\methodacro} jointly with the student. We initialize the coefficients such that the model is initialized to be the same as the \textit{static} attention explainer (i.e., performing the mean over all heads). 
\end{itemize}

Independently of the $T$-explainer, we always use a learned attention-based explainer as the $S$-explainer, considering all heads except when the $T$-explainer is a static attention explainer that only considers the last layers' heads, where we do the same for the $S$-explainer. We use the Kullback-Leibler divergence as $\mathcal{L}_{\text{expl}}$, and we set $\beta=5$ for attention-based explainers and $\beta=0.2$ for gradient-based explainers (since we found smaller values to be better). We set $\mathcal{L}_{\text{sim}}$ as the cross-entropy loss for classification tasks, and as the mean squared error loss for text regression.
For each setting, we train five students with different seeds.   Since there is some variance in students' performance (we hypothesize due to the small training sets) we report the \textbf{median} and \textbf{interquantile range (IQR)} around it (relative to the 25-75 percentile).  

\subsection{Text Classification}
\label{sec:exp:textclass}

For text classification, we consider the IMDB dataset \citep{maas-EtAl:2011:ACL-HLT2011}, a binary sentiment classification task over highly polarized English movie reviews. As the base pretrained model, we use the small ELECTRA model \citep{clark2020electra}, with 12 layers and 4 heads in each (total 48 heads). 

Like the setting in \cite{pruthi2020-evaluating}, we use the original training set with 25,000 samples to train the teacher, and further split the test set into a training set for the student and a dev and test set. We vary the number of samples the student is trained on between 500, 1,000, and 2,000. We evaluate \emph{simulability} using accuracy (i.e., what percentage of student predictions match with teacher predictions). The teacher model obtains 91\% accuracy on the student test set. 

\begin{table}[!t]\centering
\caption{\label{table:imdb-results} Results for the IMDB dataset with respect to student \textit{simulability} in terms of accuracy (\%). \textit{Underlined} values indicate higher simulability than baseline with non-overlapping IQR.}
\resizebox{0.75\textwidth}{!}{%
\begin{tabular}{lccc}\toprule
&500 &1,000 &2,000\\\midrule
No Explainer &81.72 {\footnotesize  [81.24:81.75]} &83.44 {\footnotesize   [83.36:83.63]} &84.84 {\footnotesize   [84.80:84.88]}  \\
\midrule

Gradient L2 & 81.66 {\footnotesize [81.32:82.00]} & 82.98 {\footnotesize   [82.72:83.08]} & 84.78 {\footnotesize  [84.96:85.08]} \\
Gradient $\times$ Input &\underline{84.83} {\footnotesize   [84.79:84.88]} &81.15 {\footnotesize   [80.95:81.36]} &83.84 {\footnotesize   [83.59:84.99]} \\
Integrated Gradients &\underline{82.99} {\footnotesize   [82.59:82.99]} &81.79 {\footnotesize   [81.72:81.87]} &84.20 {\footnotesize   [84.03:85.03]} \\
Attention (\textit{all layers}) & \underline{83.00} {\footnotesize   [82.60:83.00]} &\underline{85.72} {\footnotesize   [85.72:86.23]} &\underline{90.08} {\footnotesize   [89.72:90.11]} \\
Attention (\textit{last layer}) & 80.91 {\footnotesize   [79.99:81.07]} &	83.15 {\footnotesize   [82.91:83.51]} &	\underline{91.47} {\footnotesize   [91.39:91.56]} \\
\midrule
Attention (\textbf{\methodacro}) &\underline{\textbf{91.48}} {\footnotesize   [91.40:91.56]} &\underline{\textbf{92.56}} {\footnotesize   [92.28:92.83]} &\underline{\textbf{92.84}} {\footnotesize   [92.84:93.08]} \\
\bottomrule
\end{tabular}
}
\vspace{-0.5em}
\end{table}

\autoref{table:imdb-results} shows the results in terms of simulability (\autoref{eq:sim}) for the three settings. We can see that, overall, the attention explainer trained with \methodacro leads to students that simulate the teacher model much more accurately than students trained without any explanations, and more accurately than students trained with any \textit{static} explainer across all student training set sizes. Interestingly, the gradient-based explainers only improve over the baseline students when the amount of training data is very low, and actually degrade simulability for larger amounts of data (see discussion in \ref{app:explainer-details}). Using only heads from the last layer seems to have the opposite effect, leading to higher simulability than all other static explainers only for larger training sets.

\begin{minipage}[!t]{\textwidth}
\ \newline
\begin{minipage}[b]{0.74\textwidth}
    \centering
    \includegraphics[height=3.2cm]{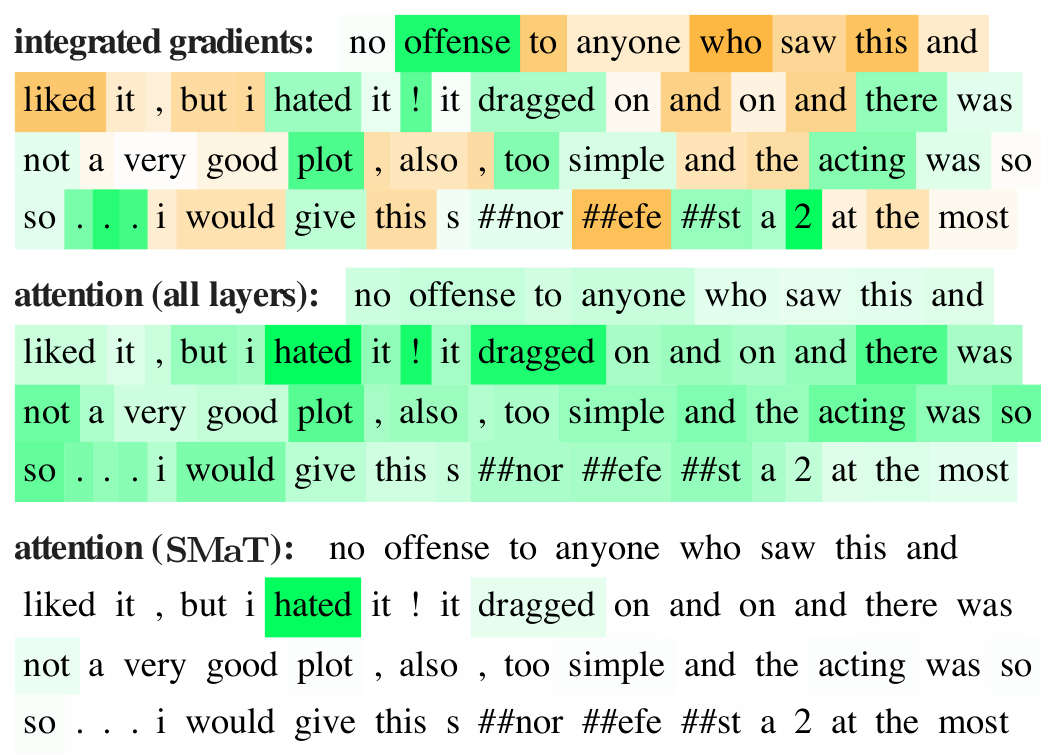}
    \includegraphics[height=3.2cm]{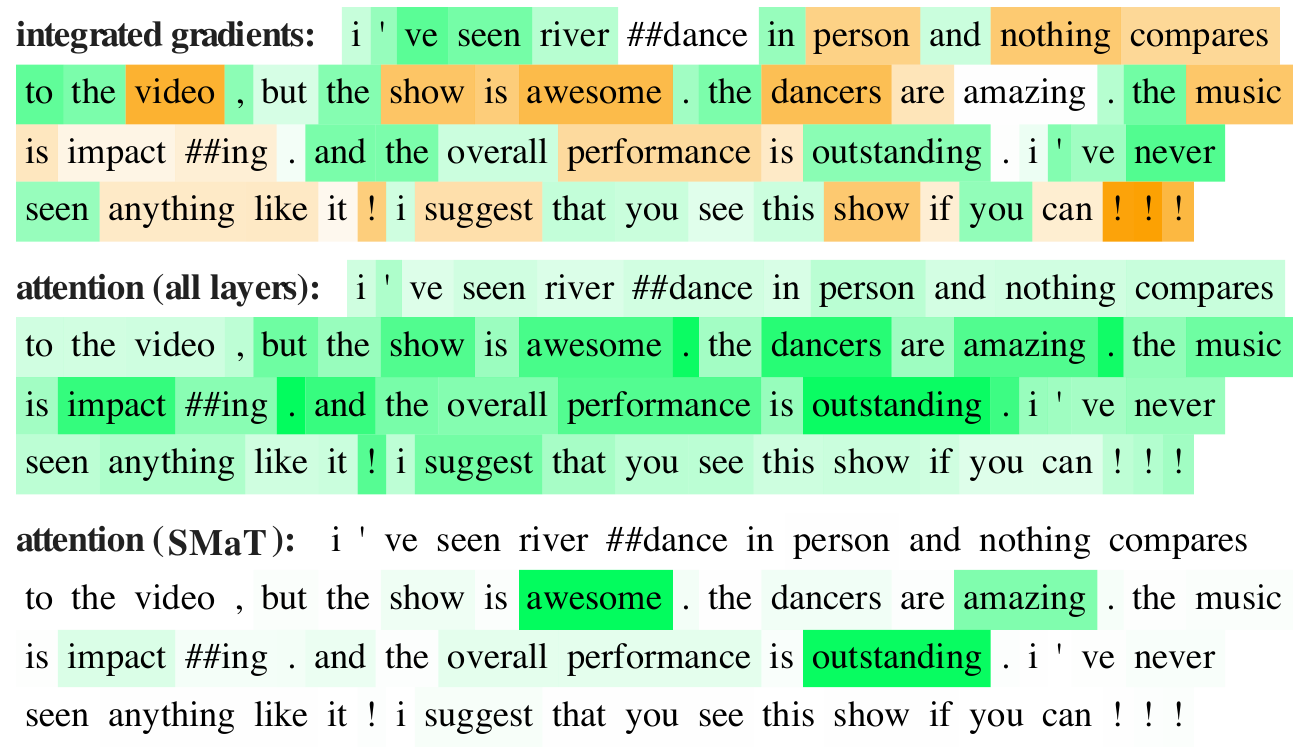}
    \captionof{figure}{Explanations given by integrated gradients, attention (\textit{last layer}), and our learned attention explainer (\methodacro) for two movie reviews of the IMDB dataset (negative and positive examples). \textcolor{green!80!black}{Green} and \textcolor{orange}{orange} represent positive and negative contributions, respectively. 
    }
    \label{fig:examples_explanations_imdb}
\end{minipage}
\hfill
\begin{minipage}[b]{0.23\textwidth}
    \centering
    \captionof{table}{\textit{Plausibility} on \textit{MovieReviews} in terms of AUC. * represents methods that use human labels.
    } \label{table:imdb-plausibility}
    \small
    \resizebox{0.8\textwidth}{!}{%
    \begin{tabular}{l@{\ }c}
        \toprule
        & AUC  \\
        \midrule
        Grad. L2 & 0.65 \\
        Grad. $\times$ Input & 0.51 \\
        Integrated Grad. & 0.53 \\
        Attn. (\textit{all layers}) & 0.68 \\
        Attn. (\textit{last layer}) & 0.61 \\
        Attn. (\textbf{\methodacro}) & \bf 0.73 \\
        \midrule
        Attn. (\textit{best layer})* & 0.75 \\
        Attn. (\textit{best head})* & 0.75 \\
        \bottomrule
    \end{tabular}
    }
    \vspace{0.5em}

\end{minipage}
\newline
\end{minipage}

\textbf{Plausibility analysis.} We select the median model trained with 1,000 samples and extract explanations for test samples from the MovieReviews dataset \citep{deyoung-etal-2020-eraser}, which contains binary sentiment movie reviews from Rotten Tomatoes %
alongside human-rationale annotation. Since the labels are binary (indicating whether a token is part of the explanation or not) and the predicted scores are real values, we follow~\cite{fomicheva-etal-2021-eval4nlp} and report our results in terms of the Area Under the Curve (AUC), which automatically considers multiple binarization thresholds. The results are shown in \autoref{table:imdb-plausibility} along with two randomly selected examples of extracted explanations in \autoref{fig:examples_explanations_imdb}. 
We found that gradient-based explanations are less plausible than those using attention (with the exception of \textit{Grad. L2}, which is similar to static attention) and that ones produced with \methodacro achieve the highest plausibility, indicating that our learned explainer can produce human-like explanations while maximizing simulability. 
Moreover, \methodacro achieves a similar AUC score to the best performing attention layer and head,\footnote{AUC scores obtained by independently trying all attention heads and layers of the model.} while not requiring \textit{any} human annotations. This is evidence that scaffolded simulability, while not explicitly designed for it, is a good proxy for plausibility and ``human-like'' explanations.\looseness=-1

\subsection{Image Classification}
\label{sec:exp:imageclass}

To validate our framework across multiple modalities, we consider image classification on the CIFAR-100 dataset \citep{Krizhevsky09learningmultiple}. We use as the base model the Vision Transformer (ViT) \citep{dosovitskiy2020image}, in particular the base version with $16\times 16$ patches that was only pretrained on ImageNet-21k \citep{Ridnik2021ImageNet21KPF}. We up-sample images to  to a $224 \times 224$ resolution.

Since the self-attention mechanism in the ViT model only works with patch representations, the explanations produced by attention-based explainers will be at patch-level rather than pixel-level. We split the original CIFAR-100 training set into a new training set with 45,000 and a validation set with 5,000. Unlike the previous task, we reuse the training set for both the teacher and student, varying the number of samples the student is trained with between 2,250 (5\%), 4,500 (10\%) and 9,000 (20\%). We use accuracy as the simulability metric and the teacher obtains 89\% on test set.

\looseness=-1
\autoref{table:cifar100-results} shows the results for the three settings. Similarly to the results in the text modality, the attention explainer trained with \methodacro achieves the best scaffolding performance, although the gaps to static attention-based explainers are smaller (especially when students are trained with more samples). Here, the gradient-based explainers always degrade simulability across the tested training set sizes and and it seems important that the explanations include attention information from layers other than the last one.

\begin{table}[!t]\centering
\caption{\label{table:cifar100-results} \textit{Simulability} results, in terms of accuracy (\%), on the CIFAR100 dataset. \textit{Underlined} values represent better performance than baseline with non-overlapping IQR}
\resizebox{0.75\textwidth}{!}{%
\begin{tabular}{lcccc}\toprule
&2,250 &4,500 &9,000 \\\midrule
No Explainer &81.16 {\footnotesize   [80.98:81.26]} &84.02 {\footnotesize   [83.98:84.24]} &85.20 {\footnotesize   [85.17:85.26]} \\
\midrule

Gradient L2 &80.97 {\footnotesize  [80.91:81.10]} &83.98 {\footnotesize [83.81:84.23]} &85.13 {\footnotesize [84.97:85.50]} \\
Gradient $\times$ Input &80.93 {\footnotesize   [80.82:81.04]} &83.99 {\footnotesize   [83.98:84.13]} &85.33 {\footnotesize   [84.85:85.35]} \\
Integrated gradients &80.22 {\footnotesize   [80.17:80.35]} &83.44 {\footnotesize   [83.25:83.44]} &84.99 {\footnotesize   [84.76:85.22]} \\
Attention (\textit{all layers}) &\underline{82.53} {\footnotesize   [82.53:82.62]} &\underline{84.81} {\footnotesize   [84.74:84.92]} &\underline{\textbf{85.92}} {\footnotesize   [85.78:85.94]} \\
Attention (\textit{last layer}) &\underline{82.34} {\footnotesize   [82.30:82.60]} &\underline{84.65} {\footnotesize   [84.56:84.81]} &85.31 {\footnotesize   [84.84:85.31]} \\
\midrule
Attention (\textbf{\methodacro}) &\underline{\textbf{83.09}} {\footnotesize   [82.77:83.28]} &\underline{\textbf{85.42}} {\footnotesize   [85.39:85.85]} &\underline{\textbf{85.96}} {\footnotesize   [85.74:86.35]} \\
\bottomrule
\end{tabular}
}
\end{table}

\textbf{Plausibility analysis.} Since there are no available human annotations for plausibility in the CIFAR-100 dataset, we design a user study to measure the plausability of the considered methods. The original image and explanations extracted with Gradient $\times$ Input, Integrated Gradients, Attention (\textit{all layers}), and Attention (\methodacro) are shown to the user, and the user has to rank the different explanations to answer the question \textit{``Which explanation aligns the most with how you would explain a similar decision?''}. Explanations were annotated by three volunteers. After collecting results, we compute the \textit{rank} and the \textit{TrueSkill} rating \citep{herbrich2007trueskill} for each explainer (roughly, the ``skill'' level if the explainers where players in game). Further description can be found in \autoref{app:visual-study}. 
The results are shown in \autoref{table:vit-plausibility}. As in the previous task, attention trained with \methodacro outperforms all other explainers in terms of plausibility, and its predicted \textit{rating} is much higher than all other explainers.
We also show examples of explanations for a set of randomly selected images in \autoref{fig:examples_explanations_vit}. 

\begin{minipage}{\textwidth}
\ \\
\begin{minipage}[b]{0.74\textwidth}
    
    \noindent 
    \begin{tabular}[l]{@{}l@{\ }l@{\ }l@{\ }l@{} c@{} l@{\ }l@{\ }l@{\ }l}
      \scriptsize{Input image} & 
      \scriptsize{Integ. Grad.} &
      \scriptsize{Attn. (all lx.)} &
      \scriptsize{Attn. (\methodacro)} & \qquad &
      \scriptsize{\ Input image} & 
      \scriptsize{Integ. Grad.} &
      \scriptsize{Attn. (all lx.)} &
      \scriptsize{Attn. (\methodacro)}
    \end{tabular}
    \centering
    \\
    \includegraphics[width=0.115\textwidth]{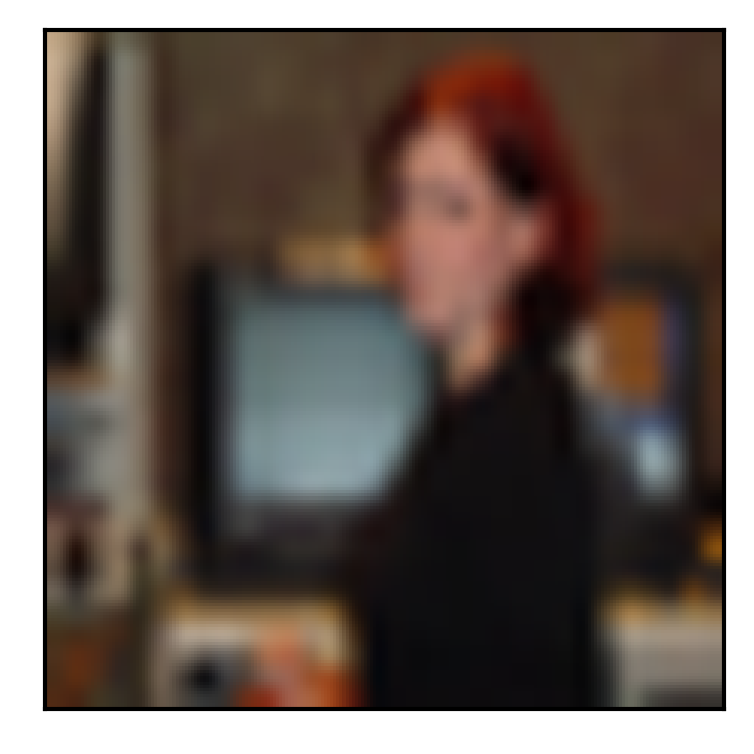}
    \includegraphics[width=0.115\textwidth]{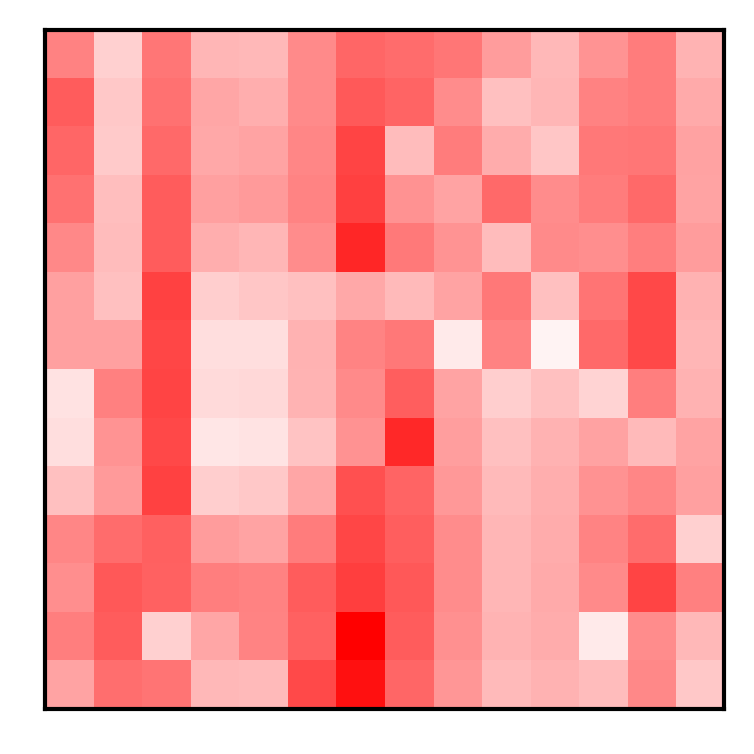}
    \includegraphics[width=0.115\textwidth]{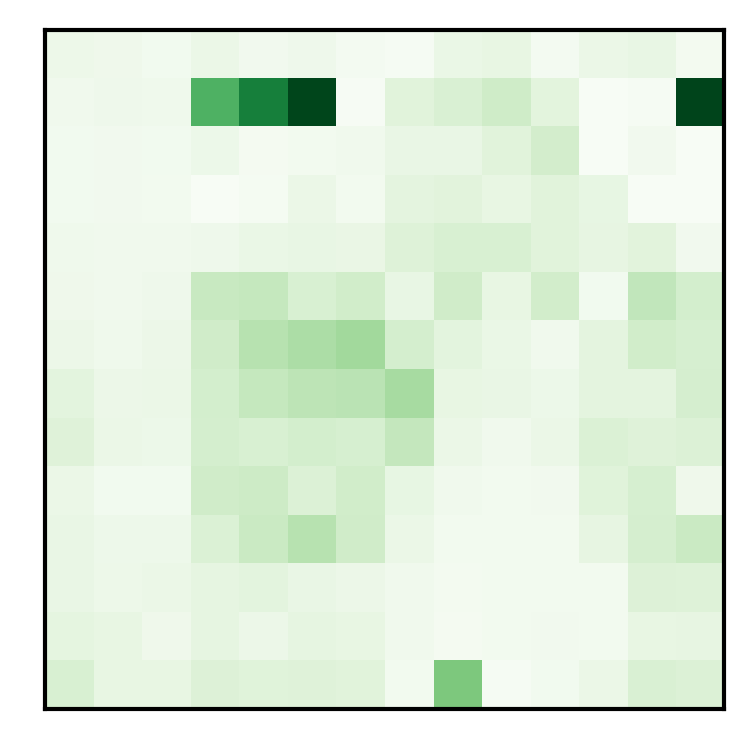}
    \includegraphics[width=0.115\textwidth]{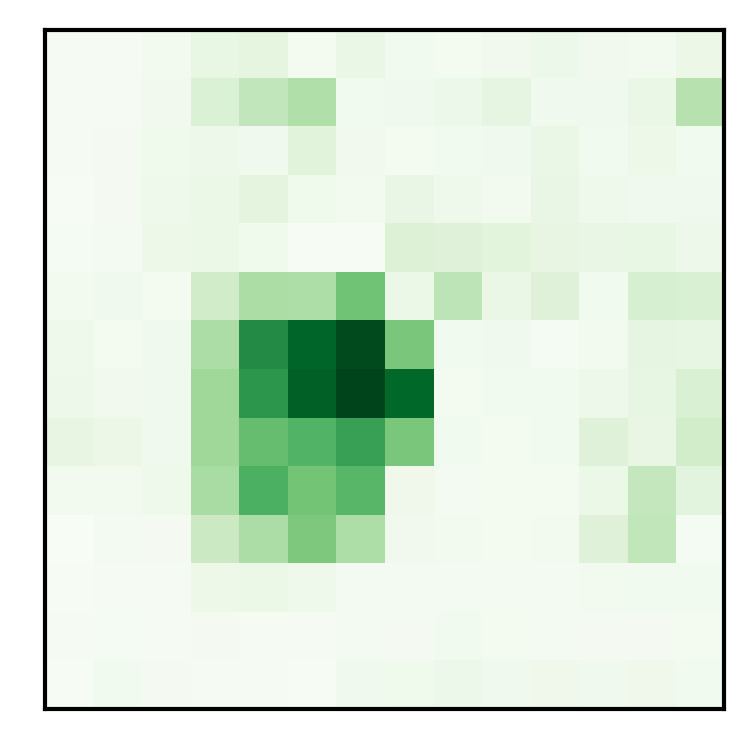}
    \hfill
    \includegraphics[width=0.115\textwidth]{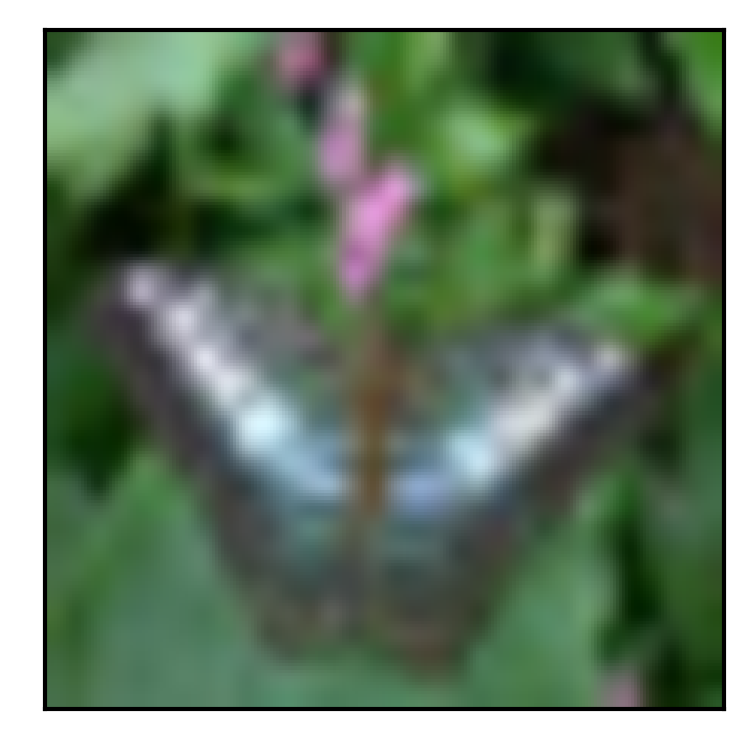}
    \includegraphics[width=0.115\textwidth]{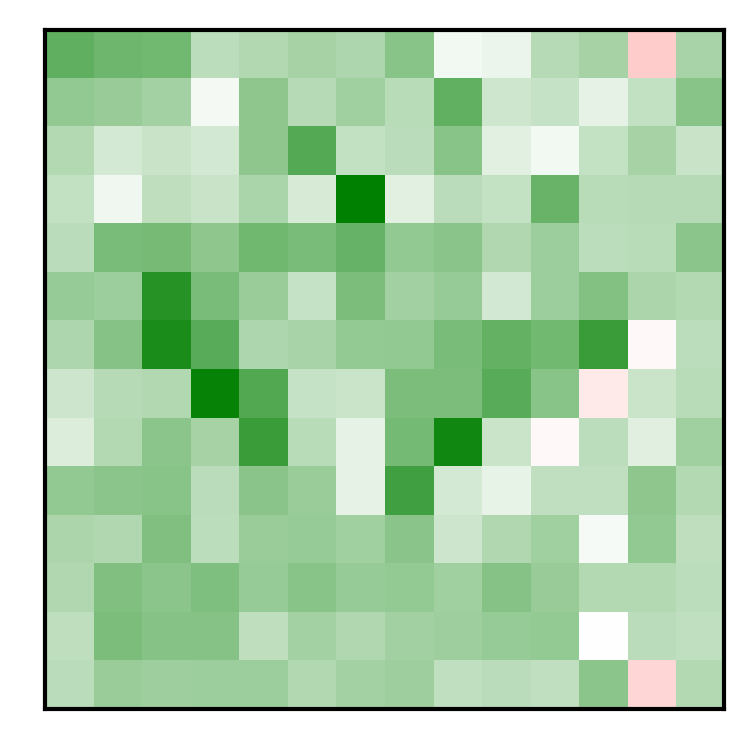}
    \includegraphics[width=0.115\textwidth]{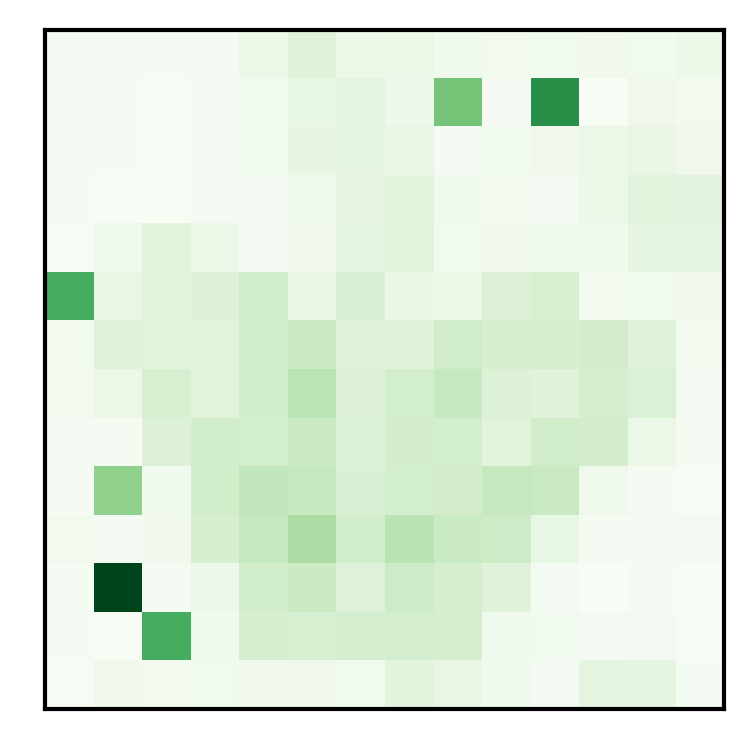}
    \includegraphics[width=0.115\textwidth]{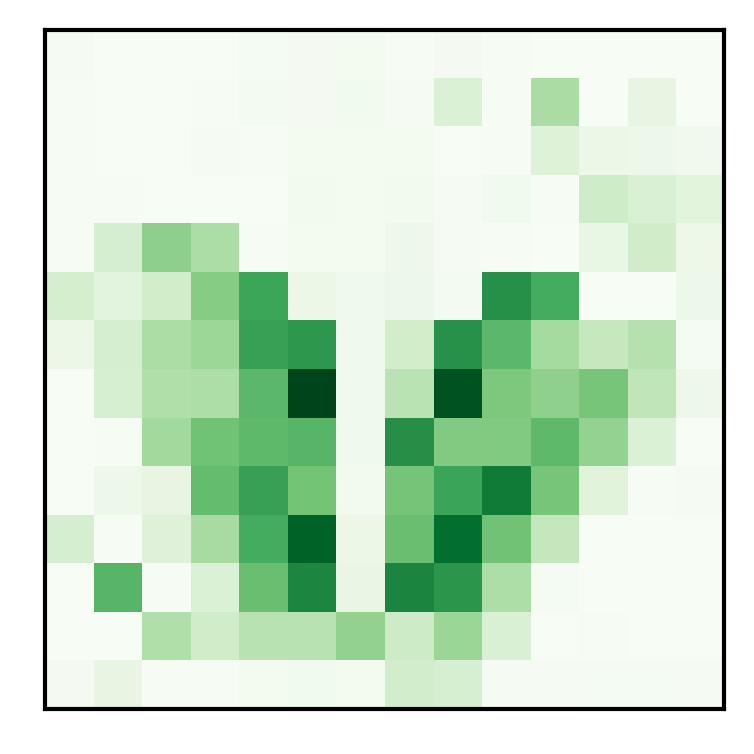}
    \\
    \includegraphics[width=0.115\textwidth]{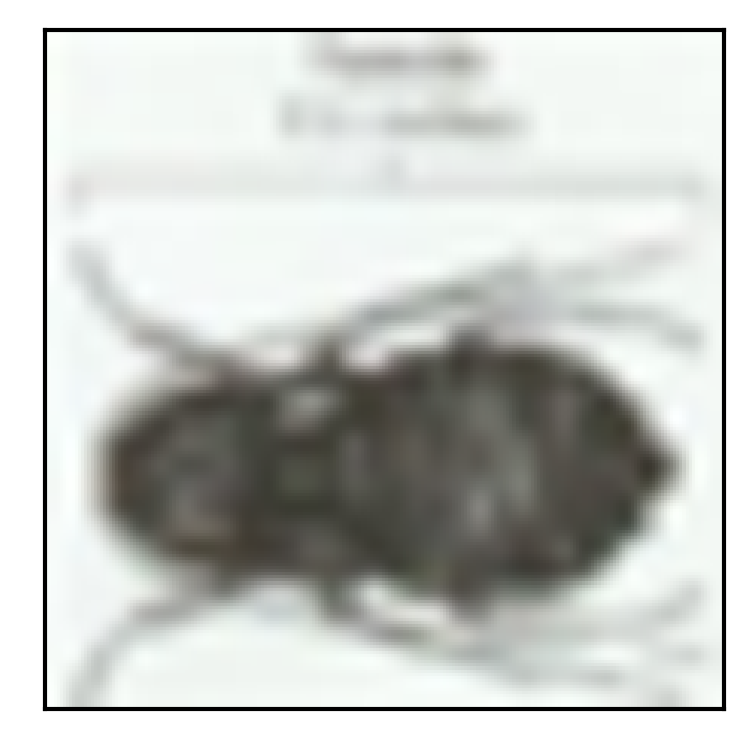}
    \includegraphics[width=0.115\textwidth]{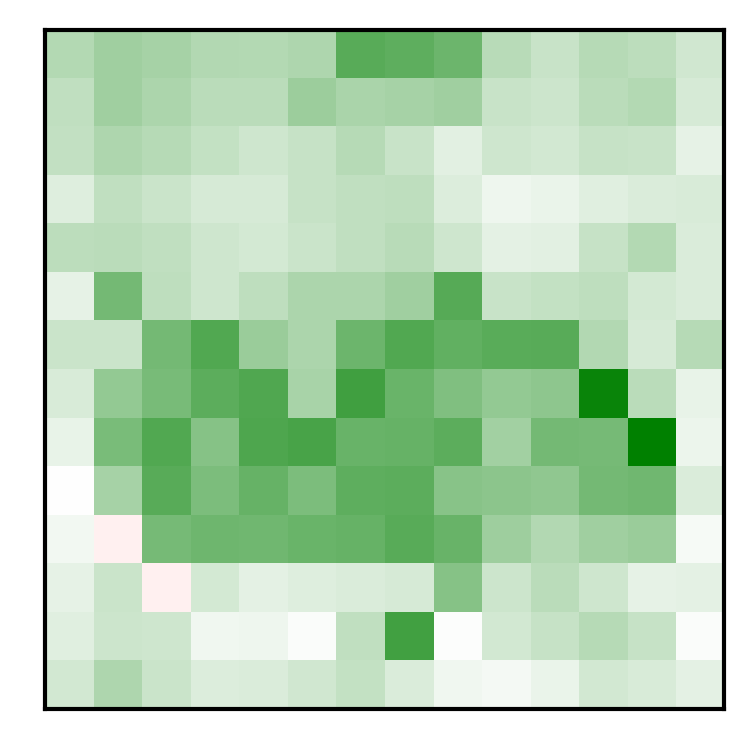}
    \includegraphics[width=0.115\textwidth]{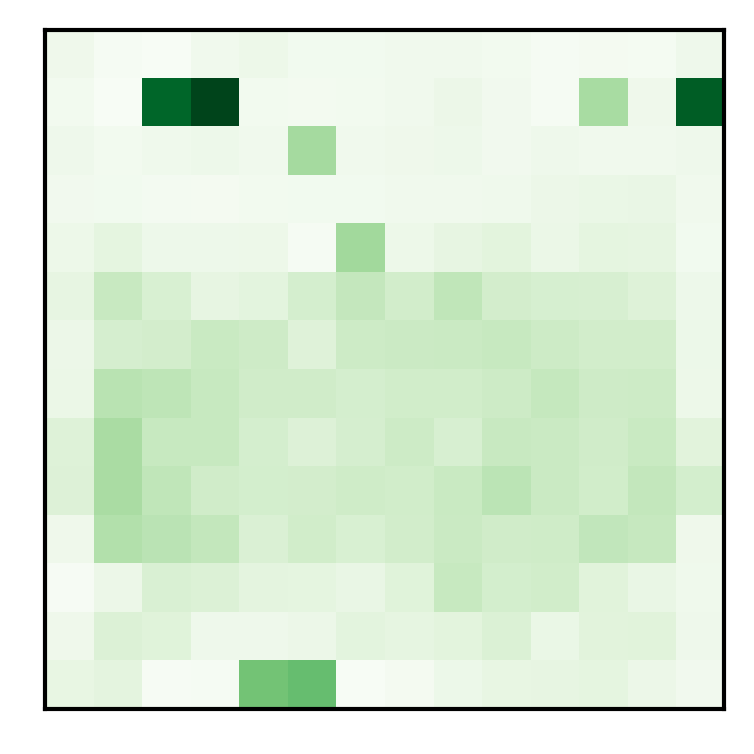}
    \includegraphics[width=0.115\textwidth]{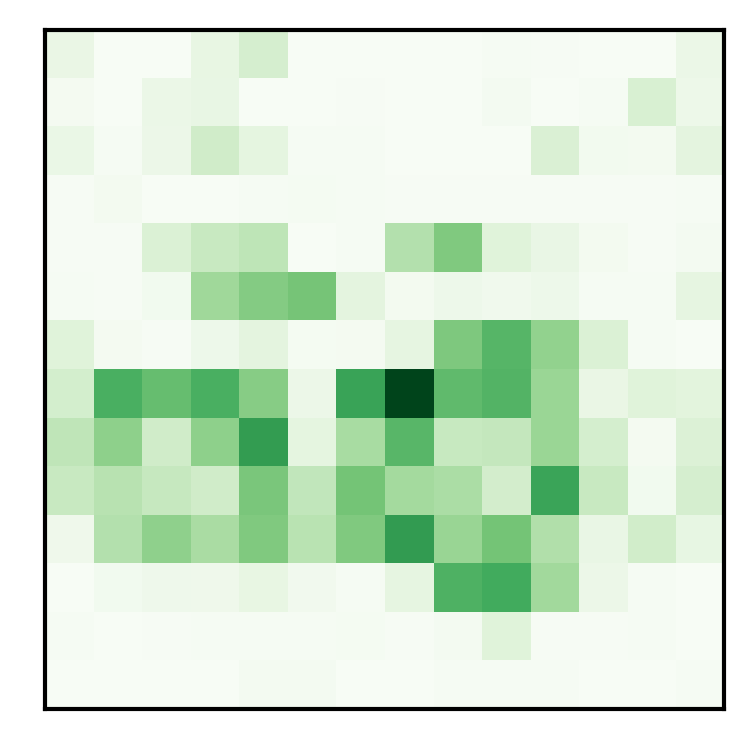}
    \hfill
    \includegraphics[width=0.115\textwidth]{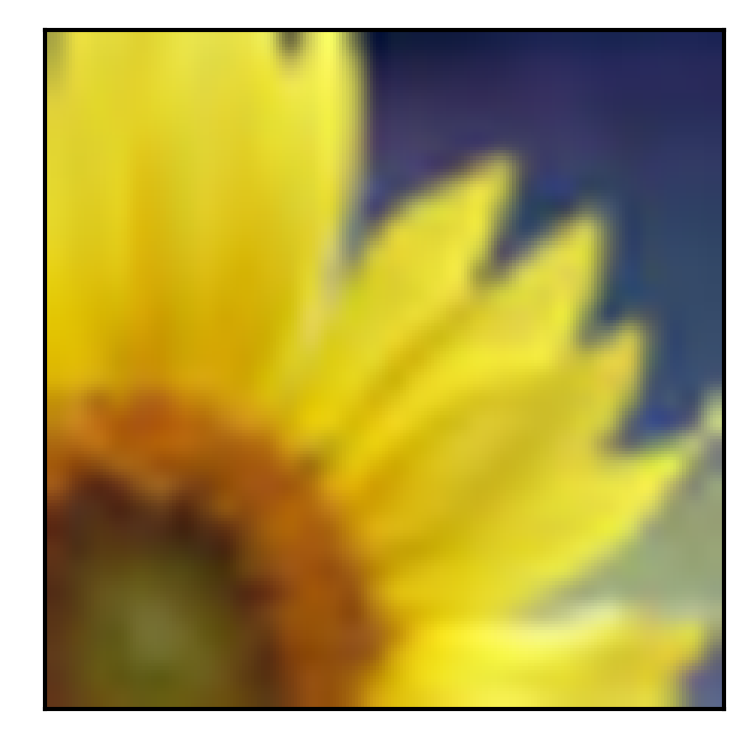}
    \includegraphics[width=0.115\textwidth]{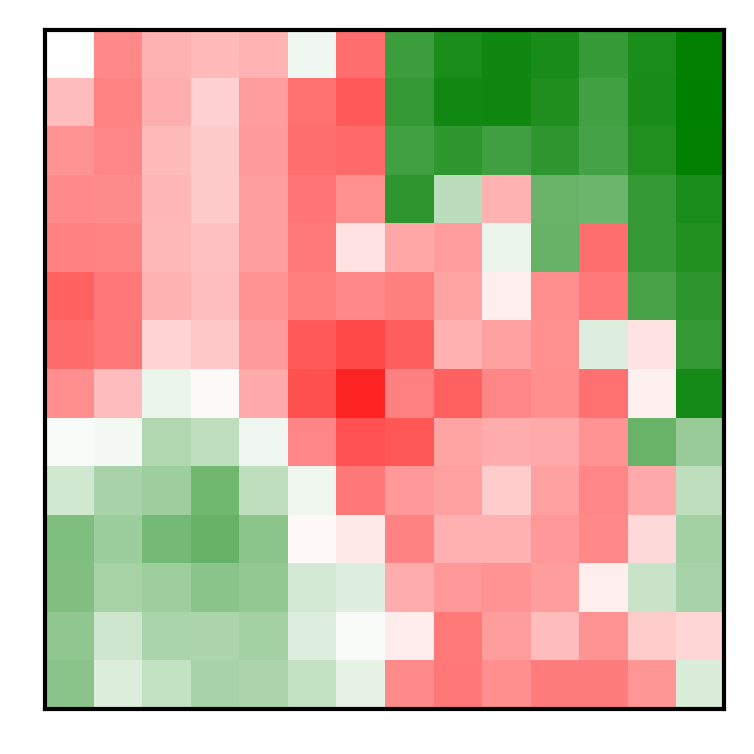}
    \includegraphics[width=0.115\textwidth]{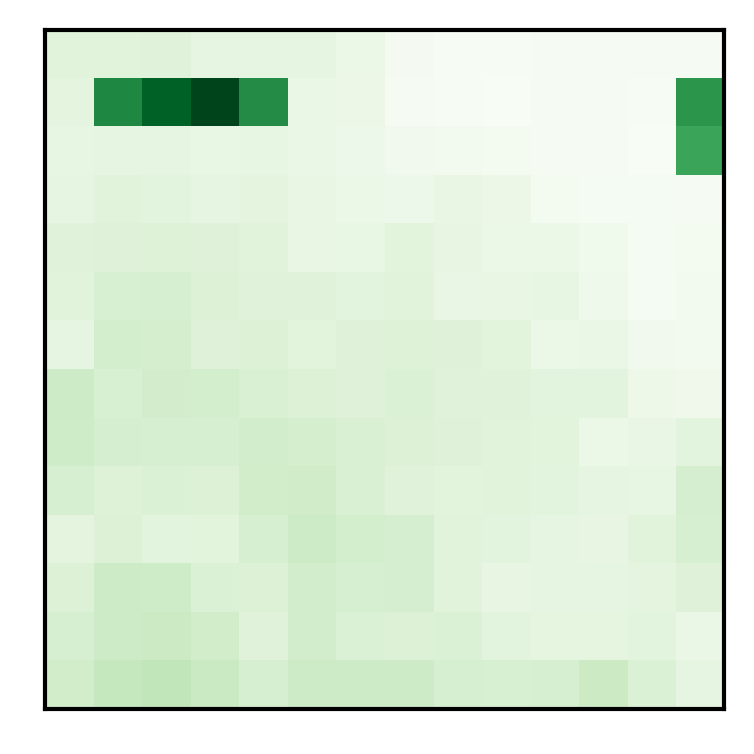}
    \includegraphics[width=0.115\textwidth]{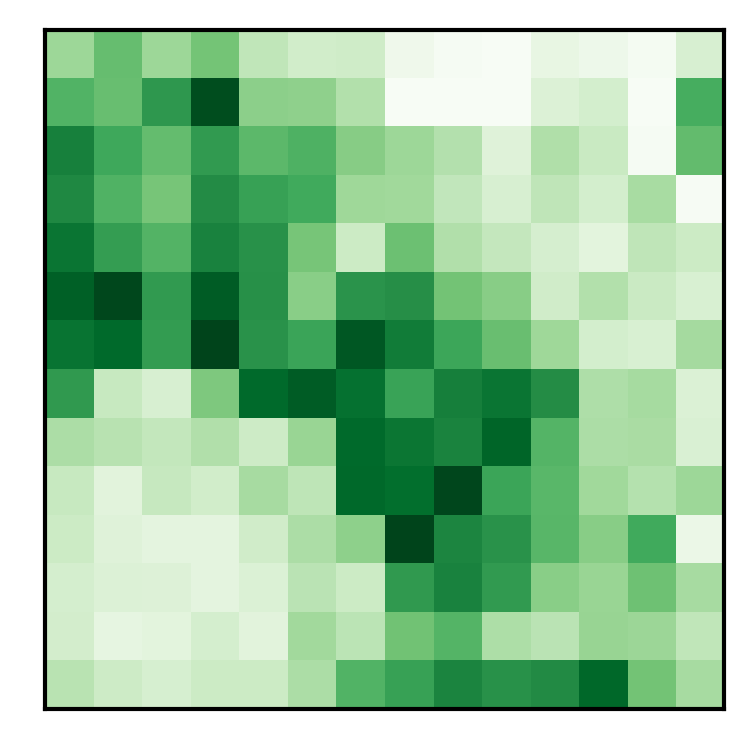}

    \captionof{figure}{Explanations given by integrated gradients, attention (\textit{all layers}), and learned attention explainer for a set of input images of CIFAR-100. Gold labels are: ``television'', ``butterfly'', ``cockroach'', and ``sunflower''.}
    \label{fig:examples_explanations_vit}
\end{minipage}
\hfill
\begin{minipage}[b]{0.23\textwidth}
    \centering
    \captionof{table}{\textit{Plausibility} results of the human study on visual explanations.
}\label{table:vit-plausibility}
    \resizebox{1\textwidth}{!}{%
    \setlength{\tabcolsep}{2pt}
    \renewcommand{\arraystretch}{1.1}
    \begin{tabular}{l@{\ }cr}
        \toprule
        & Rank & \textit{TrueSkill}  \\
        \midrule
        Grad. $\times$ Input & 3-4 & -2.7±.67 \\
        Integ. Grad. & 3-4 & -2.1±.67 \\
        Attn. (\textit{all lx.}) & 2 & 0.7±.67 \\
        Attn. (\textbf{\methodacro}) & \textbf{1} & \textbf{4.3±.70} \\
        \bottomrule
    \end{tabular}
    }
    \vspace{0.8em}

\end{minipage}
\vspace{-0.5em}
\newline
\end{minipage}

\subsection{Machine Translation Quality Estimation}
\label{sec:exp:qe}

Quality Estimation (QE) is the task of predicting a quality score given a sentence in a source language and a translation in a target language from a machine translation system, which requires models that consider interactions between the two inputs, source and target. Scores tend to be continuous values (making this a regression task) that were collected from expert annotators.

Interpreting quality scores of machine translated outputs is a problem that has received recent interest \citep{fomicheva-etal-2021-eval4nlp} since it allows identifying which words were responsible for a bad translation.
We use the MLQE-PE dataset \citep{tacl2020}, which contains 7,000 training samples for each of seven language pairs alongside word-level human annotation. We use as the base model a pretrained XLM-R-base \citep{conneau2019unsupervised}, a multilingual model with 12 layers and 12 heads in each (total of 144 heads).

We exclude one of the language pairs in the dataset (\texttt{si-en}) since the XLM-R model did not support it, leading to a training set with 42,000 samples. Similar to the CIFAR-100 case, we reuse the same training set for both the teacher and student, sampling a subset for the latter. We vary the number of samples the student is trained with between 2,100 (5\%), 4,200 (10\%) and 8,400 (20\%). Since this is a regression task, we evaluate simulability using the Pearson correlation coefficient between student and teacher's predictions.\footnote{Pearson correlation is the standard metric used to evaluate sentence-level QE models.} The teacher achieves 0.63 correlation on the test set.

\begin{table}[!t]\centering
\caption{\label{table:mlqe-results} \textit{Simulability} results, in terms of Pearson correlation, on the ML-QE dataset. \textit{Underlined} values represent better performance than baseline with non-overlapping IQR.}
\resizebox{0.8\textwidth}{!}{%
\begin{tabular}{lcccc}\toprule
&2,100 &4,200 &8,400 \\\midrule
No Explainer &.7457 {\footnotesize   [.7366:.7528]} &.7719 {\footnotesize   [.7660:.7802]} &.7891 {\footnotesize   [.7860:.7964]} \\
\midrule

Gradient L2 & \underline{.8065} {\footnotesize   [.8038:.8268]} & \underline{.8535} {\footnotesize   [.7117:.8544]} & \underline{\textbf{.8638}} {\footnotesize   [.8411:.8657] } \\
Gradient $\times$ Input &.6846 {\footnotesize   [.6781:.6894]} &.6922 {\footnotesize   [.6885:.6965]} &.7141 {\footnotesize   [.7136:.7147]} \\
Integrated gradients &.6686 {\footnotesize   [.6677:.6694]} &.7086 {\footnotesize   [.6994:.7101]} &.7036 {\footnotesize   [.6976:.7037]} \\
Attention (\textit{all layers}) &\underline{.8120} {\footnotesize   [.7955:.8125]} &\underline{.8193} {\footnotesize   [.8186:.8280]} &\underline{\textbf{.8467}} {\footnotesize   [.8464:.8521]} \\
Attention (\textit{last layer}) &.7486 {\footnotesize   [.7484:.7534]} &.7720 {\footnotesize   [.7672:.7726]} &.7798 {\footnotesize   [.7717:.7814]} \\
\midrule
Attention (\textbf{\methodacro}) &\underline{\textbf{.8156}} {\footnotesize   [.8096:.8183]} &\underline{\textbf{.8630}} {\footnotesize   [.8412:.8724]} &\underline{\textbf{.8561}} {\footnotesize   [.8512:.8689]} \\
\bottomrule
\end{tabular}
}
\vspace{-0.5em}
\end{table}

\autoref{table:mlqe-results} shows the results for the three settings. Similar to other tasks, the attention explainer trained with \methodacro leads to students with higher simulability than baseline students and similar or higher than \textit{static} explainer across all training set sizes. Curiously, the \textit{Grad. L2} explainer achieves very high simulability for this task. It even has a higher \textit{median} simulability score than \methodacro for 8,400 samples. However, we attribute this to variance in the student training set sampling (that could lead to an imbalance in language pair proportions) which could explain why \methodacro performance degrades with more samples. For this task, the gradient-based explainers always degrade simulability across the tested training set size. It also seems that using only the last layer's attention is also ineffective at teaching students, achieving the same performance as the baseline.

\begin{table}[!t]
    \caption{\label{tab:plausibility_mlqe}Plausibility results for source and target inputs for each language pair of the MLQE-PE dataset in terms of AUC. * represents \textit{supervised} methods that use human labels in some form. }
    \centering
    \small
    \resizebox{\textwidth}{!}{%
    \begin{tabular}{l c@{\hskip 0.2cm} c@{\hskip 0.2cm}c@{\hskip 0.2cm}c@{\hskip 0.2cm} c@{\hskip 0.2cm}c@{\hskip 0.2cm}c@{\hskip 0.2cm} c@{\hskip 0.2cm}c@{\hskip 0.2cm}c@{\hskip 0.2cm} c@{\hskip 0.2cm}c@{\hskip 0.2cm}c@{\hskip 0.2cm} c@{\hskip 0.2cm}c@{\hskip 0.2cm}c@{\hskip 0.2cm} c@{\hskip 0.2cm}c@{\hskip 0.2cm}c@{\hskip 0.2cm} c@{\hskip 0.2cm}c@{\hskip 0.2cm}}
        \toprule 
        & & 
        \multicolumn{2}{c}{\sc en-de} & & 
        \multicolumn{2}{c}{\sc en-zh} & & 
        \multicolumn{2}{c}{\sc et-en} & & 
        \multicolumn{2}{c}{\sc ne-en} & & 
        \multicolumn{2}{c}{\sc ro-en} & & 
        \multicolumn{2}{c}{\sc ru-en} & & 
        \multicolumn{2}{c}{\sc overall} \\
        \cmidrule{3-4}
        \cmidrule{6-7}
        \cmidrule{9-10}
        \cmidrule{12-13}
        \cmidrule{15-16}
        \cmidrule{18-19}
        \cmidrule{21-22}
        & & src. & tgt. & & src. & tgt. & & src. & tgt. & & src. & tgt. & & src. & tgt. & & src. & tgt. & & src. & tgt. \\
        \midrule 
        Gradient L2 & & \bf 0.64 & \bf 0.65 & & 0.65 & 0.49 & & \bf 0.67 & 0.61 & & \bf 0.68 & \bf 0.55 & & \bf 0.72 & 0.68 & & \bf 0.65 & 0.54 & & \bf 0.67 & 0.59 \\

        Gradient $\times$ Input & & 0.58 & 0.60 & & 0.61 & 0.51 & & 0.60 & 0.54 & & 0.61 & 0.49 & & 0.64 & 0.59 & & 0.58 & 0.51 & & 0.61 & 0.54 \\
        
        Integrated Gradients & & 0.59 & 0.60 & & 0.63 & 0.49 & & 0.60 & 0.52 & & 0.64 & 0.48 & & 0.64 & 0.59 & & 0.60 & 0.51 & & 0.62 & 0.53 \\
        
        Attention (\textit{all layers}) & & 0.60 & 0.63 & & \textbf{0.68} & \textbf{0.52} & & 0.60 & 0.61 & & 0.58 & \bf 0.55 & & 0.66 & \bf 0.70 & & 0.62 & \bf 0.55 & & 0.62 & 0.59 \\
        Attention (\textit{last layer}) & & 0.51 & 0.49 & & 0.61 & 0.49 & & 0.51 & 0.50 & & 0.55 & 0.48 & & 0.52 & 0.57 & & 0.56 & 0.50 & & 0.54 & 0.50 \\
        Attention (\textbf{\methodacro}) & & \bf 0.64 & \bf 0.65 & & \textbf{0.68} & \textbf{0.52} & & 0.66 & \bf 0.64 & & 0.66 & 0.54 & & 0.71 & \bf 0.70 & & 0.61 & 0.54 & & 0.66 & \bf 0.60 \\
        \midrule
        Attention (\textit{best layer})* & & 0.64 & 0.65 & & 0.69 & 0.64 & & 0.64 & 0.68 & & 0.68 & 0.68 & & 0.71 & 0.76 & & 0.64 & 0.59 & & 0.65 & 0.65 \\
        Attention (\textit{best head})* & & 0.67 & 0.67 & & 0.70 & 0.65 & & 0.70 & 0.70 & & 0.70 & 0.69 & & 0.73 & 0.75 & & 0.67 & 0.60 & & 0.67 & 0.66 \\
        \bottomrule
    \end{tabular}
    }
    \vspace{-0.5em}
\end{table}

\textbf{Plausibility analysis.} We select the median model trained with 4,200 samples and follow the approach devised in the Explainable QE shared task to evaluate plausibility  \citep{fomicheva-etal-2021-eval4nlp}, which consists of evaluating the human-likeness of explanations 
in terms of AUC only on the subset of translations that contain errors. The results are shown in \autoref{tab:plausibility_mlqe}. We note that for all language pairs, \methodacro performs on par or better than static explainers, and only being surpassed by \textit{Grad. L2} in  the \textit{source-side} over all languages. Comparing with the best attention layer/head, an approach used by \citet{fomicheva2021translation,treviso-etal-2021-ist}, \methodacro achieves similar AUC scores for source explanations, but lags behind the best attention layer/head for target explanations on \textsc{*-en} language pairs. However, as stressed previously for text and image classification, \methodacro sidesteps human annotation and avoids the cumbersome approach of independently computing plausibility scores for all heads. \looseness=-1

\section{Importance of the Head Projection}
\label{app:importance-head-projection}

A major component of our framework is the normalization of the head coefficients, as defined in \autoref{sec:attention-explainer}. Although many functions can be used to map scores to probabilities, we found empirically that \textsc{Sparsemax} performs the best, while other transformations such as \textsc{Softmax} and \textsc{1.5-Entmax} \citep{peters-etal-2019-sparse}, a sparse transformation more dense than sparsemax, usually lead to poorly performing students (see \autoref{table:normalization}). 

\begin{table}[!htp]

\caption{\label{table:normalization} \textit{Simulability} results, in terms of accuracy (\%), on the MLQE dataset with 4200 training examples, with different normalization functions. 
}
\centering

\resizebox{0.95\textwidth}{!}{%
\begin{tabular}{lcccc}\toprule
&\textsc{Sparsemax} & \textsc{Softmax} & \textsc{1.5-Entmax} & No Normalization\\\midrule
No Explainer &.7719 {\footnotesize ±  [.7660:.7802]}&.7719 {\footnotesize ±  [.7660:.7802]}&.7719 {\footnotesize ±  [.7660:.7802]}&.7719 {\footnotesize ±  [.7660:.7802]}\\
\midrule
Attention (\textit{all layers})  &\underline{.8193} {\footnotesize ±  [.8186:.8280]} & .7345{\footnotesize ± [.7335:.7390]} &.7152 {\footnotesize ± [.7111:.7161]} &.7781 {\footnotesize ± [.7762:.7791]} \\
Attention (\textit{last layer}) &.7720 {\footnotesize ±  [.7672:.7726]} &.7697{\footnotesize ± [.7659:.7715]} &.7807 {\footnotesize ± [.7652:.7821]} &.7768 {\footnotesize ± [.7764:.7807]} \\
\midrule
Attention (\textbf{\methodacro}) &\underline{\textbf{.8630}} {\footnotesize ±  [.8412:.8724]} &.7439 {\footnotesize ± [.7430:.7484]} &.7163 {\footnotesize ± [.7130:.7239]} & \underline{.8002} {\footnotesize ± [.7919:.8100]} \\
\bottomrule
\end{tabular}
}
\end{table}

\begin{figure}[!htb]
    \centering
    \includegraphics[height=4cm]{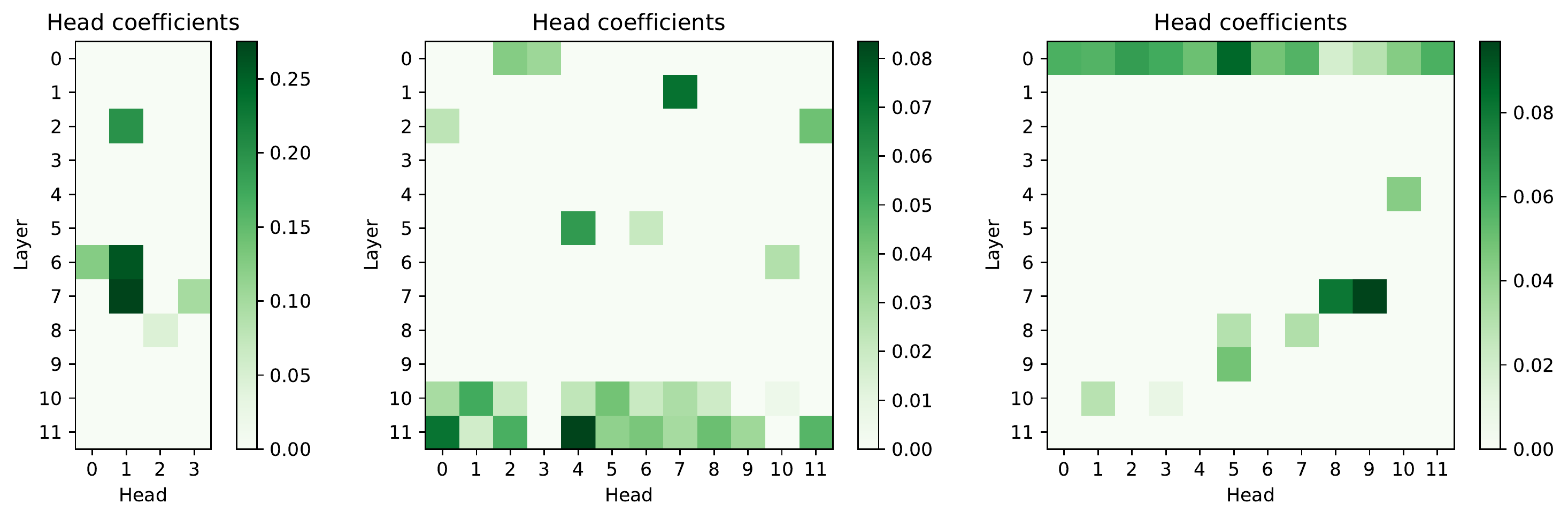}
    \vspace{-0.7em}
    \caption{Head coefficients for text classification (left), image classification (middle), and quality estimation (right), illustrating that only a small subset of attention heads are deemed relevant by SMaT due to \textsc{Sparsemax}.
    }
    \label{fig:head_coeffs}
\end{figure}

Furthermore, another benefit of \textsc{Sparsemax} is that it produces a small subset of \textit{active} heads. The heatmaps of attention coefficients ($\lambda_T$) learned after training, shown in \autoref{fig:head_coeffs}, exemplify this. We can see that the dependency between head position (layer it belongs to) and its coefficient is task/dataset/model specific, with CIFAR-100 and MLQE having opposite observations. We also found empirically that \textit{active} heads ($\lambda_T^h > 0$) usually lead to higher plausibility scores than \textit{inactive} heads, further reinforcing the good plausibility findings of \methodacro.

\section{Related Work}
\label{sec:related-work}

\textbf{Explainability for text \& vision.}
Several works propose explainability methods to interpret decisions made by NLP and CV models. Besides gradient and attention-based approaches already mentioned, some extract explanations by running the models with perturbed inputs \citep{lime,feng-etal-2018-pathologies,kim-etal-2020-interpretation}. Others even define custom backward passes to assign relevance for each feature \citep{Bach2015OnPE}. These methods are commonly employed together with post-processing heuristics, such as selecting only the top-k tokens/pixels with higher scores for visualization. Another line of work seeks to build a classifier with inherently interpretable components, such as methods based on attention mechanisms and rationalizers \citep{lei2016rationalizing,bastings-etal-2019-interpretable}.

\textbf{Evaluation of explainability methods.} As mentioned in the introduction, early works evaluated explanations based on properties such as \textit{consistency}, \textit{sufficiency} and \textit{comprehensiveness}. 
\citet{jacovi-goldberg-2020-towards} recommended the use of a graded notion of faithfulness, which the ERASER benchmark quantifies using the idea of sufficient and comprehensive rationales, alongside compiling datasets with human-annotated rationales for calculating plausibility metrics \citep{deyoung-etal-2020-eraser}. Given the disagreement between explainability methods, \citet{neely2021order} showed that without a faithful ground-truth explanation it is impossible to determine which method is better. Diagnostic tests such as the ones proposed by \citet{adebayo2018sanity,wiegreffe-pinter-2019-attention} and \citet{atanasova-etal-2020-diagnostic} are more informative yet they do not capture the main goal of an explanation: the ability to communicate an explanation to a practitioner.

\textbf{Simulability.} A new dimension for evaluating explainability methods relies on the forward prediction/simulation proposed by \citet{lipton-2018-mythos} and \citet{doshi2017towards}, which states that humans should be able to correctly simulate the model's output given the input and the explanation. 
\citet{chandrasekaran-etal-2018-explanations,hase-bansal-2020-evaluating, arora22aaai} analyze simulability via human studies across text classification datasets.
\citet{treviso-martins-2020-explanation} designed an automatic framework where students (machine or human) have to predict the model's output given an explanation as input. 
Similarly, \citet{pruthi2020-evaluating} proposed the simulability framework that was extended in our work, where explanations are used to regularize the student rather than passed as input. 

\textbf{Learning to explain.} 
The concept of simulability also opens a path to learning explainers.  In particular \citet{treviso-martins-2020-explanation} learn an attention-based explainer that maximizes simulability. However, directly optimizing for simulability sometimes led to explainers that learned trivial protocols (such as selecting only punctuation symbols or stopwords to leak the label).
Our approach of optimizing a teacher-student framework is similar to approaches that optimize for model distillation \citep{zhou2021meta}. However, these approaches modify the original model rather than introduce a new explainer module. \citet{raghu2020teaching} propose a framework similar to ours for learning \textit{commentaries} for inputs that speed up and improve the training of a model. However commentaries are model-independent and are optimised to improve performance on the real task.
Rationalizers \citep{chen2018learning,jacovi-yoav-2021-aligning,guerreiro-martins-2021-spectra} also directly learn to extract explanations, but can also suffer from trivial protocols. 

\section{Conclusion \& Future Work}
\label{sec:conclusion}

We proposed \textbf{\methodacro}, a framework for directly optimizing explanations of the model's predictions to improve the training of a student \textit{simulating} the said model. We found that, across tasks and domains, explanations learned with SMaT both lead to students that simulate the original model more accurately and are more aligned with how people explain similar decisions when compared to previously proposed methods. On top of that, our parameterized attention explainer provides a principled way for discovering relevant attention heads in transformers.

Our work shows that scaffolding is a suitable criterion for both evaluating and optimizing explainability methods, and we hope that SMaT paves way for new research to develop expressive interpretable components for neural networks that can be directly trained without  human-labeled explanations. However, %
it should be noted that ``interpretability'' is %
a loosely defined concept, 
and therefore caution should be exercised when making statements about the quality of explanations based only on simulability, especially if these statements might have societal impacts. 

We only explored learning attention-based explainers, but our method can also be used to optimize other types of explainability methods, including gradient-based ones, by introducing learnable parameters in their formulations. Another promising future research direction is to explore using \methodacro to learn explanations other than saliency maps.

\section*{Acknowledgments}
This work was supported by the European Research Council (ERC StG DeepSPIN 758969), by EU's Horizon Europe Research and Innovation Actions 
(UTTER, contract 101070631), 
by P2020 project MAIA (LISBOA-01-0247- FEDER045909), and Fundação para a Ciência e Tecnologia through project PTDC/CCI-INF/4703/2021 (PRELUNA) and contract UIDB/50008/2020. We are grateful to Nuno Sabino, Thales Bertaglia, Henrico Brum, and Antonio Farinhas for the participation in human evaluation experiments.

\bibliographystyle{plainnat}
\bibliography{main}

\section*{Checklist}

\begin{enumerate}

\item For all authors...
\begin{enumerate}
  \item Do the main claims made in the abstract and introduction accurately reflect the paper's contributions and scope? Yes
  \item Did you describe the limitations of your work?
  Yes, see \autoref{sec:conclusion}
  \item Did you discuss any potential negative societal impacts of your work?
  Yes, see \autoref{sec:conclusion}
  \item Have you read the ethics review guidelines and ensured that your paper conforms to them?
  Yes
\end{enumerate}

\item If you are including theoretical results...
\begin{enumerate}
  \item Did you state the full set of assumptions of all theoretical results?
  N/A
  \item Did you include complete proofs of all theoretical results?
  N/A
\end{enumerate}

\item If you ran experiments...
\begin{enumerate}
  \item Did you include the code, data, and instructions needed to reproduce the main experimental results (either in the supplemental material or as a URL)?
  Yes
  \item Did you specify all the training details (e.g., data splits, hyperparameters, how they were chosen)?
  Yes, with unspecified ones included in the code
  \item Did you report error bars (e.g., with respect to the random seed after running experiments multiple times)?
  Yes
  \item Did you include the total amount of compute and the type of resources used (e.g., type of GPUs, internal cluster, or cloud provider)?
   No
\end{enumerate}

\item If you are using existing assets (e.g., code, data, models) or curating/releasing new assets...
\begin{enumerate}
  \item If your work uses existing assets, did you cite the creators?
  Yes
  \item Did you mention the license of the assets?
  No
  \item Did you include any new assets either in the supplemental material or as a URL?
  Yes
  \item Did you discuss whether and how consent was obtained from people whose data you're using/curating?
  No
  \item Did you discuss whether the data you are using/curating contains personally identifiable information or offensive content?
  No
\end{enumerate}

\item If you used crowdsourcing or conducted research with human subjects...
\begin{enumerate}
  \item Did you include the full text of instructions given to participants and screenshots, if applicable?
  Yes, see \autoref{app:visual-study}
  \item Did you describe any potential participant risks, with links to Institutional Review Board (IRB) approvals, if applicable?
  N/A
  \item Did you include the estimated hourly wage paid to participants and the total amount spent on participant compensation?
  N/A (Volunteers)
\end{enumerate}

\end{enumerate}

\appendix

\section{Explainer Details}
\label{app:explainer-details}

With the \textit{integrated gradients} explainer \citep{pmlr-v70-sundararajan17a}, we use 10 iterations for the integral in the \textit{simulability} experiments (due to the computation costs) and 50 iterations for the \textit{plausability} experiments. We use zero vectors as baseline embeddings, since we found little variation in changing this. For both gradients-based explainers, we project into the simplex by using the \textsc{Softmax} function, similar to the attention-based explainers. This results in very negative values having low probability values. Moreover, for evaluating plausibility on text classification and translation quality estimation, we computed the explanation score of a single word by summing the scores of its word pieces.

We would like to note that, unlike the setting in \cite{pruthi2020-evaluating}, we \textbf{do not} apply a \textit{top-k} post-processing heuristic on gradients/attention logits, instead directly projecting them to the simplex. This might explain the difference in results to the original paper, particularly for the low simulability performance of static explainers.

\section{Human Study for Visual Explanations}
\label{app:visual-study}

\begin{figure}[h]
    \centering
    \includegraphics[width=0.3\textwidth]{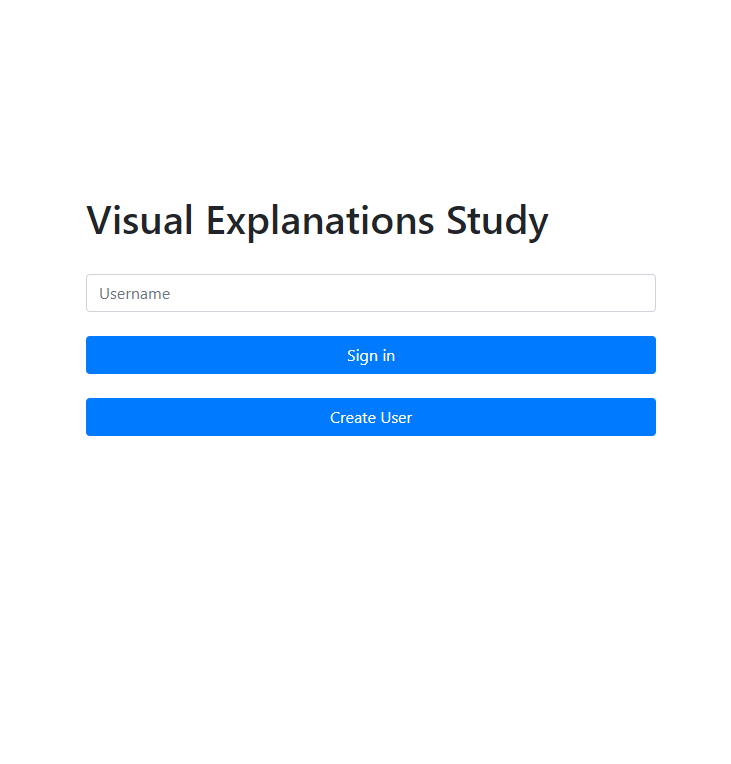}
    \includegraphics[width=0.25\textwidth]{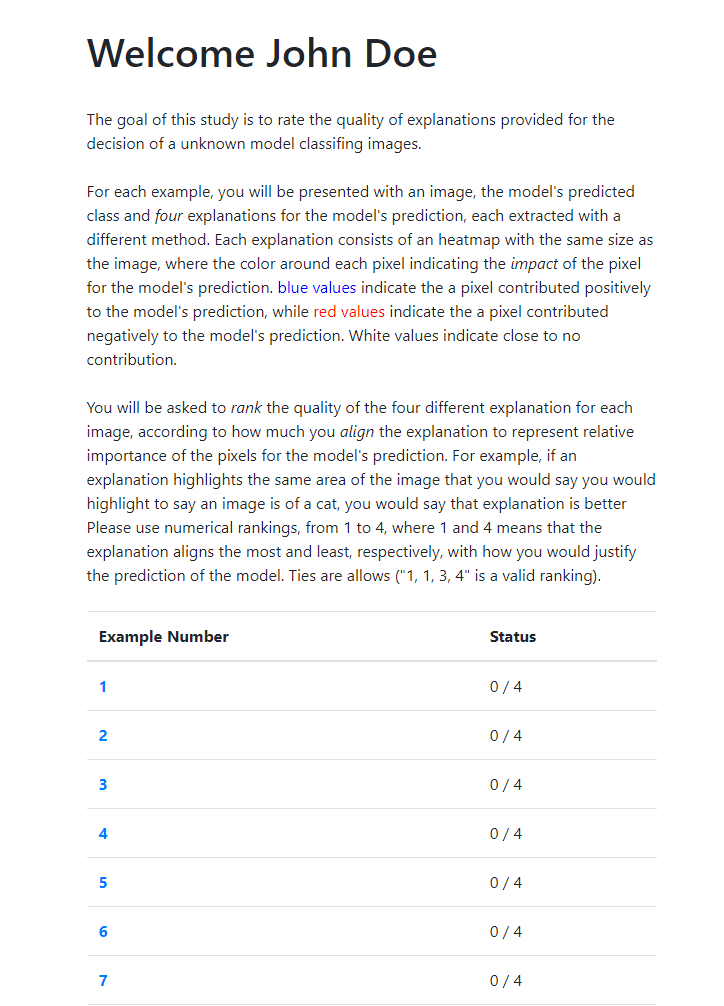}
    \includegraphics[width=0.35\textwidth]{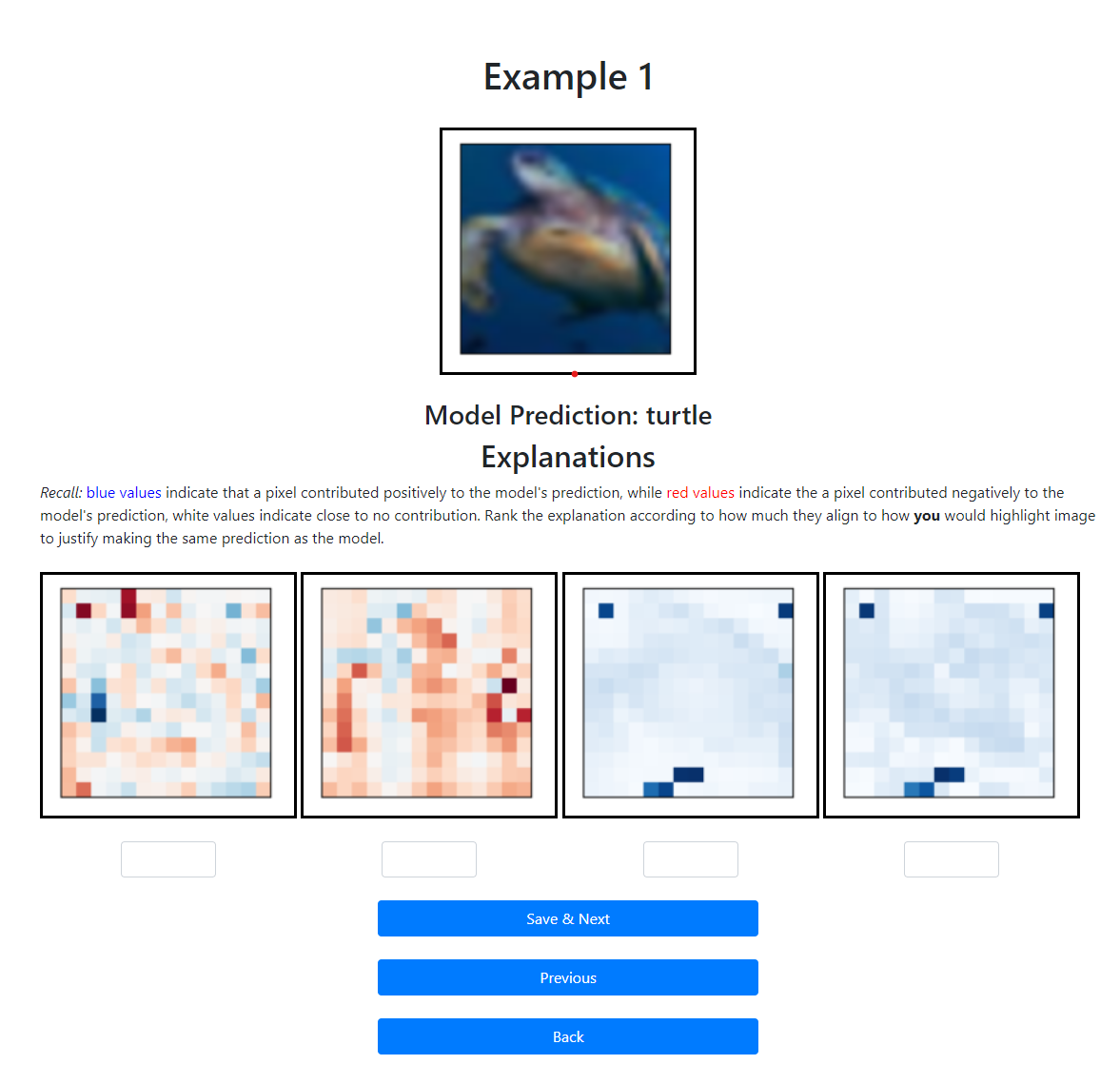}
    \caption{Login page (left), dashboard (middle) and annotation page (right)}
    \label{fig:webpage}
\end{figure}

The annotations were collected through an annotation webpage, built on top of Flask. \autoref{fig:webpage} shows the three pages of the site. During the annotation, users were asked to rank four explanations, unnamed and in random order. After collecting the ratings, we computed the \textit{TrueSkill} rating, with an initial rating for each method of $\mu=0, \sigma=0.5$. After learning the ratings, we then compute the \textit{ranks} by obtaining the 95\% confidence interval for the rating each method, and constructing a partial ordering of methods based on this. 

The volunteers were a mixture of graduates or graduate students known by the authors. However we would like to point out that due to blind nature of the method annotation, the chance of bias is low.

\begin{figure}[h]
    \centering
    Original image (``television'')
    \\
    \includegraphics[width=0.07\textwidth]{figs/examples-vit/5-tv/image.png}
    \\
    \includegraphics[width=0.85\textwidth]{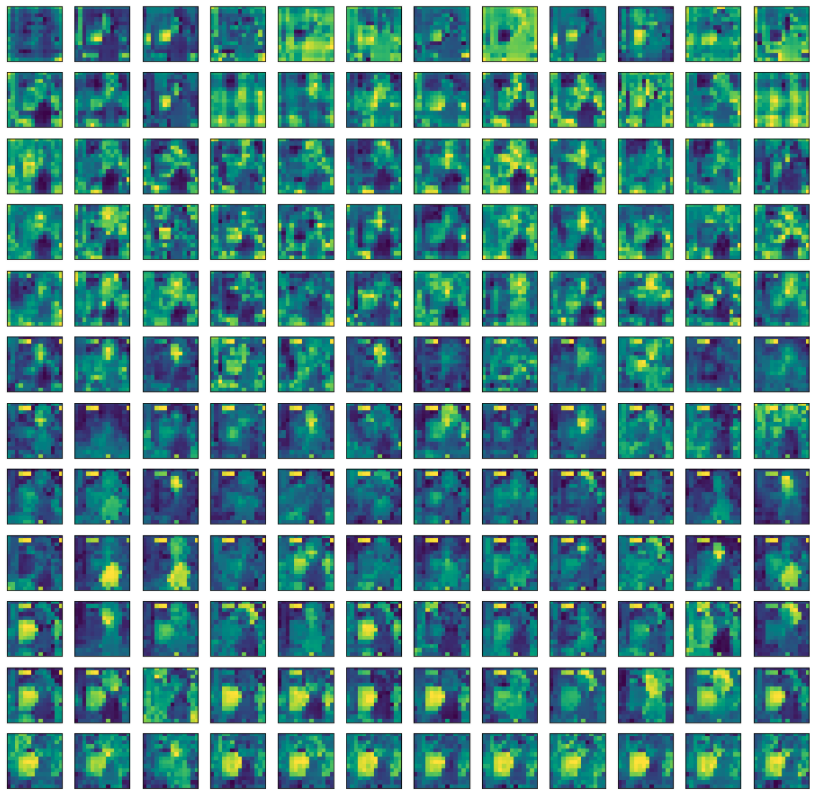}
    \caption{Explanations from all layers (rows) and heads (columns) of the CIFAR-100 teacher model.}
    \label{fig:tv_example_attn_learned_all_layers_heads}
\end{figure}

\begin{figure}[h]
    \centering
    Original image (``butterfly'')
    \\
    \includegraphics[width=0.07\textwidth]{figs/examples-vit/114-butterfly/image.png}
    \\
    \includegraphics[width=0.85\textwidth]{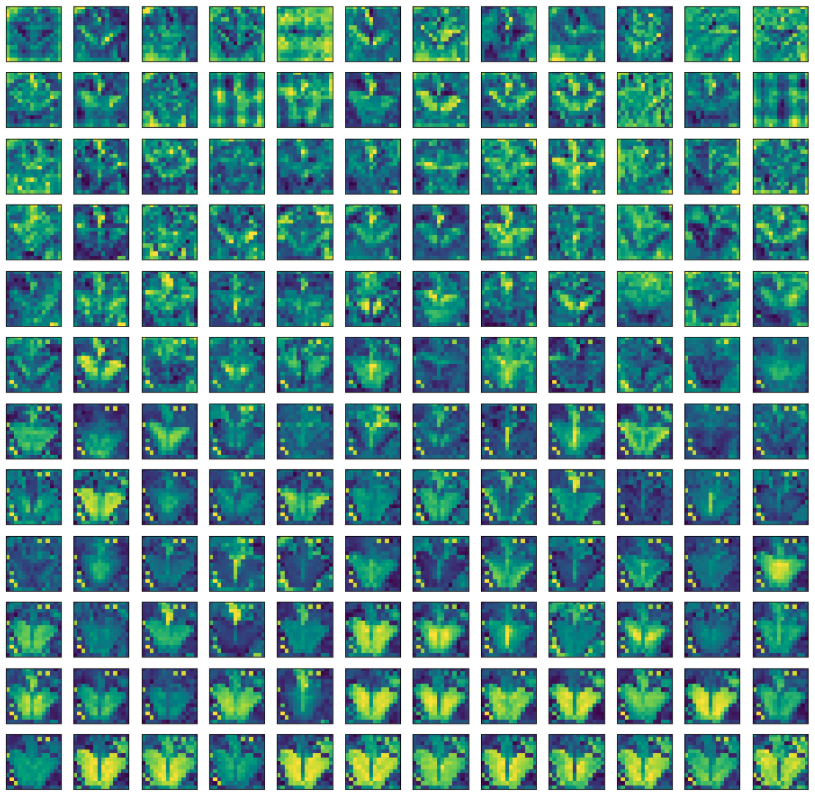}
    \caption{Explanations from all layers (rows) and heads (columns) of the CIFAR-100 teacher model.}
    \label{fig:butterfly_example_attn_learned_all_layers_heads}
\end{figure}

In Figures~\ref{fig:tv_example_attn_learned_all_layers_heads} and \ref{fig:butterfly_example_attn_learned_all_layers_heads} we show raw attention explanations extracted from all layers (rows) and heads (columns) of the teacher transformer used in our CIFAR-100 experiments. Cross-checking the explanations with the most relevant heads selected by SMaT for image classification, we can see that most selected heads produce plausible explanations (e.g., attention heads from the last layers).

\end{document}